\documentclass[a4paper,fleqn]{cas-sc}

\usepackage{amssymb}
\usepackage{amsmath}
\usepackage{graphics}
\usepackage{graphicx}
\usepackage{bm}
\usepackage{caption, subcaption}

\usepackage{cleveref}
\usepackage{dirtytalk}

\newcommand{\f}{
f(x_t, h_{t-1}; w_h)
}
\newcommand{\g}{
g(h_{t}; w_o)
}
\renewcommand{\l}{
\mathcal{L}
}

\newcommand{\fwh}{
\frac{
\partial \f
}{
\partial w_h
}
}

\newcommand{\fh}{
\frac{
\partial \f
}{
\partial h_{t-1}
}
}

\newcommand{\fx}{
\frac{
\partial \f
}{
\partial x_t
}
}

\newcommand{\hw}{
\frac{
\partial h_t
}{
\partial w_h
}
}

\newcommand{\xh}{
\frac{
\partial g(h_{t-1})
}{
\partial h_{t-1}
}
}

\newcommand{\hmw}{
\frac{
\partial h_{t-1}
}{
\partial w_h
}
}

\newcommand{\xw}{
\frac{
\partial x_t
}{
\partial w_h
}
}

\newcommand{\Real}{\mathbb{R}}

\definecolor{blue}{rgb}{0,0,1} 
\definecolor{green}{rgb}{0.0, 0.5020, 0.0} 
\definecolor{red}{rgb}{1,0,0} 
\definecolor{black}{rgb}{0,0,0} 
\definecolor{blueviolet}{rgb}{0.5412,0.1686,0.8863} 
\definecolor{orange}{rgb}{1,0.6471,0} 
\definecolor{cornflowerblue}{rgb}{0.3922,0.5843,0.9294}

\usepackage{float}
\newtheorem{lemma}{Lemma}

\usepackage[authoryear]{natbib}

\begin{document}
\let\WriteBookmarks\relax
\def\floatpagepagefraction{1}
\def\textpagefraction{.001}
\shorttitle{Robust long-term forecasting with BPTT-SA}
\shortauthors{PR Vlachas et~al.}

\title [mode = title]{
Learning from Predictions: Fusing Training and Autoregressive Inference for Long-Term Spatiotemporal Forecasts
}

\author[1,2]{PR Vlachas}[type=editor,orcid=0000-0002-3311-2100]
\ead{pvlachas@ethz.ch}
\credit{Conceptualization of the study, methodology, software development, data curation, analysis of the results, original draft preparation}

\author[1,2]{P Koumoutsakos}[orcid=0000-0001-8337-2122]
\credit{Consultation, analysis of the results, original draft preparation}
\ead{petros@ethz.ch}

\address[1]{Computational Science and Engineering Laboratory, ETH Z\"urich, Clausiusstrasse 33, Z\"urich CH-8092, Switzerland}
\address[2]{School of Engineering and Applied Sciences, 29 Oxford Street, Harvard University, Cambridge, MA 02138, USA}

\begin{abstract}
Recurrent Neural Networks (RNNs) have become an integral part of modeling and forecasting frameworks in areas like natural language processing and high-dimensional dynamical systems such as turbulent fluid flows. To improve the accuracy of predictions, RNNs are trained using the Backpropagation Through Time (BPTT) method to minimize prediction loss. During testing, RNNs are often used in autoregressive scenarios where the output of the network is fed back into the input. However, this can lead to the exposure bias effect, as the network was trained to receive ground-truth data instead of its own predictions. This mismatch between training and testing is compounded when the state distributions are different, and the train and test losses are measured. To address this, previous studies have proposed solutions for language processing networks with probabilistic predictions. Building on these advances, we propose the Scheduled Autoregressive BPTT (BPTT-SA) algorithm for predicting complex systems. Our results show that BPTT-SA effectively reduces iterative error propagation in Convolutional RNNs and Convolutional Autoencoder RNNs, and demonstrate its capabilities in long-term prediction of high-dimensional fluid flows.
\end{abstract}



\begin{keywords}
Autoregressive forecasting \sep RNN, LSTM \sep BPTT \sep Exposure bias
\end{keywords}

\maketitle

\section{Introduction}

Deep learning (DL) methods have been instrumental to advances  in a wide range of scientific disciplines from physics~\citep{baldi2014searching, lusch2018deep}, fluid dynamics~\citep{novati2019controlled, brunton2020machine}, mathematics~\citep{han2018solving}, climate modeling~\citep{kurth2018exascale}, computer vision~\citep{gregor2015draw, krizhevsky2017imagenet}, language~\citep{mikolov2011extensions} and signal processing~\citep{oord2016wavenet} to biology~\citep{alipanahi2015predicting}, medicine and drug discovery~\citep{chen2018rise}.
The successes of DL is mostly attributed to novel architectures, whose weights (parameters) are learned with optimization algorithms such as stochastic gradient descent. In DL, the key computational bottleneck is the efficient computation of gradients with error backpropagation.

The process of Backpropagation (BP) is divided into three stages: the forward pass, the backward pass, and the update step.
During the forward pass, the network's output is calculated using the input data. 
Based on the network's output and the target, the deviation is calculated, and the loss is determined. 
In the backward pass, the gradient of the loss with respect to the network's parameters is computed by applying the chain rule, which starts from the network's output and ends at the input. 
Finally, in the update step, the gradient is used to update the network's weights in the direction of minimizing the loss, guided by an optimization algorithm such as gradient descent. 
BP addresses the credit assignment problem by determining the contribution of each neuron to the network's overall performance, and updates its value to achieve the goal encoded in the loss function that needs to be minimized.

Recurrent neural networks (RNNs) are commonly used to handle sequential or temporal data as they efficiently take into account the sequential aspect of the tasks. 
The extension of the backpropagation (BP) algorithm to RNNs and temporal tasks is known as backpropagation-through-time (BPTT)~\citep{werbos1988generalization,elman1990finding,werbos1990backpropagation}.
BPTT has found widespread application in  natural language processing~\citep{mikolov2011extensions}, signal and image processing~\citep{gregor2015draw,oord2016wavenet},  speech recognition~\citep{ahmad2004recurrent} and forecasting of complex dynamical systems ~\citep{vlachas2018data}.
It has been instrumental in solving complex temporal credit assignment problems~\citep{gers2002learning, lillicrap2019backpropagation}.
In the computer vision community, video prediction has been a recent research focus~\citep{mathieu2015deep, castrejon2019improved, fragkiadaki2015learning, walker2014patch, srivastava2015unsupervised, oh2015action}, and BPTT is often utilized in a probabilistic context, such as in variational RNNs~\citep{castrejon2019improved, chung2015recurrent}.

Several recent studies have found that RNNs can effectively model and forecast high-dimensional chaotic spatiotemporal dynamics in a deterministic setting~\citep{geneva2020modeling, vlachas2018data, vlachas2020backpropagation, wan2018data, pathak2018model, pathak2017using}. 
Moreover, RNNs coupled with Convolutional Neural Networks (CNNs), \citep{shi2015convolutional}, \citep{vlachas2022multiscale,wiewel2019latent} can be employed to model high dimensional spatiotemporal data, such as flow fields or images.
The importance of long-term prediction of fluid flows is paramount for various practical cases from prediction of extreme events~\citep{blonigan2019extreme}, traffic management~\citep{li2017diffusion}, surrogate modeling~\citep{wiewel2019latent}, typhoon alert systems, to climate and precipitation forecasting~\citep{kumar2020convcast, rasp2020weatherbench, shi2015convolutional, shi2017deep}.
A recent literature survey on long-term spatiotemporal forecasting is given in~\cite{shi2018machine}.

Most commonly, RNNs are trained using the Backpropagation Through Time (BPTT) algorithm in the teacher forcing (BPTT-TF) mode~\citep{sutskever2013training}, where the network is trained to minimize the error in one-step ahead predictions using sequences from the training dataset as input.
However, our work suggests that the weight gradients computed during training in this mode may be biased towards one-step ahead predictions.
The training loss is computed based on the probability distribution of the training data, which may not match the probability distribution of the testing data during the autoregressive testing phase, where the network uses its own predictions as input. 
This discrepancy is known as exposure bias~\citep{schmidt2019Generalization}, and it can negatively impact the generalization performance of the RNN.

To address this issue, alternative training methods such as scheduled sampling~\citep{bengio2015scheduled} or curriculum learning~\citep{bengio2009curriculum} can be employed to train the network in a way that better matches the distribution of the testing data. 
By doing so, the network can better generalize to unseen data and produce more accurate predictions.

The primary objective of these works is to address the issue of discrepancy in the teacher forcing mode by incorporating techniques to replace, mask, or alter the ground-truth context. 
In the NLP field, other studies have attempted to overcome the limitations of BPTT-TF. 
For example, some studies have employed deep auto-regressive models without hidden memory states, such as those discussed in~\cite{miller2018recurrent}, including models like PixelCNN~\citep{oord2016pixel}, WaveNet~\citep{oord2016wavenet}, and models that use attention mechanisms~\citep{gregor2014deep, vaswani2017attention}. 
These alternative approaches aim to improve the generalization performance of RNNs and overcome the issue of exposure bias.

Our proposed approach for training autoregressive deep learning models for time series forecasting is based on the scheduled sampling method discussed in~\cite{bengio2015scheduled} for probabilistic RNNs. 
However, our method allows for back-propagation through the predicted outputs and differs in that the outputs are not derived from a sampled distribution.

We propose a new technique called BPTT-SA, which incorporates an auxiliary loss that accounts for the autoregressive (iterative) forecasting error and adapts the BPTT computational graph to refine the gradient computation.
The training process follows a schedule that starts with a standard one-step ahead prediction loss and gradually switches to the autoregressive loss as training progresses.
The goal of BPTT-SA is to address exposure bias in RNNs by explicitly accounting for the dissimilarity between the probability distributions of training and testing data. 
This enables the proposed approach to enhance the model's generalization performance in autoregressive forecasting.

We evaluate the effectiveness of BPTT-SA  in deterministic spatiotemporal prediction using RNNs, and compare it to standard BPTT and the scheduled sampling approach of~\cite{bengio2015scheduled}.

We note that BPTT-TF focuses on minimizing one-step-ahead prediction error, biasing it towards short-term forecasting. 
On the other hand, the autoregressive loss prioritizes long-term error. 
Balancing these conflicting objectives is challenging, even for linear prediction models~\citep{lin1994forecasting}.
We find that the benefits of BPTT-SA are not significant for low dimensional time series prediction, as demonstrated by the Mackey-Glass system and the Darwin sea level temperature time-series datasets. 
However, in the Navier-Stokes flow past a cylinder, BPTT-SA was able to reconcile the short-term accuracy and long-term accuracy objectives more effectively compared to the scheduled sampling approach of~\cite{bengio2015scheduled}, without incurring extra training cost. 
Additionally, BPTT-SA helps alleviate the propagation of errors and enhances long-term prediction.

A recent study by Teutsch et al.~\citep{teutsch2022flipped} explored the use of training algorithms for RNNs, with a focus on similar curriculum learning methods. 
However, their investigation was limited to the application of these methods to low-order chaotic systems. 
Our work expands on these findings by demonstrating that the advantages of using curriculum learning for RNNs in forecasting tasks are even more pronounced in high-dimensional dynamic systems.

\section{Models and Methods}

Recurrent neural networks (RNNs) are deep learning architectures designed specifically to handle sequential data. 
They process an input stream ${x_0, \dots, x_T}$ sequentially, where each element in the stream is a state $x \in \Real^{d_{x}}$. 
RNNs have an internal hidden state that encodes information about the history of the input stream.
The functional form of the RNN is described by:
\begin{gather}
h_t = \f \\
o_t = \g
\label{eq:rnn}
\end{gather}
where $\f$ is the recurrent (hidden-to-hidden) mapping, and $\g$ is the output (hidden-to-output) mapping
$w_h$ and $w_o$ are trainable weights of the mappings.
Gated Recurrent Units~\citep{chung2014empirical} and Long Short-Term Memory units~\citep{hochreiter1997long} are possible implementations of the aforementioned RNN mappings.
They both can be framed, however, under this unifying lens.

In multiple tasks, RNNs are used as regressors for forecasting the evolution of the state $x_t$.
In such cases, the output is a prediction of the state at the next time-step $x_{t+1}$, i.e. 
\begin{gather}
o_t=\widetilde{x}_{t+1} = \g=w_o h_t,
\label{eq:out}
\end{gather}
where $w_o \in \Real^{d_x \times d_x}$.
The weights are optimized to minimize the prediction loss $\l(x_{t+1}, o_t)$, e.g., the mean squared error:
\begin{gather}
\l(x_{t+1}, o_t) =
(x_{t+1}-o_{t})^2 = 
(x_{t+1}-\widetilde{x}_{t+1})^2
\label{eq:loss}
\end{gather}
The loss measures the difference between the RNN prediction $o_t$ and the target value $x_{t+1}$.

\subsection{Backpropagation Through Time with Teacher Forcing}
\label{sec:sub:tf}

\begin{figure}[pos=H]
\centering
\begin{subfigure}[tbhp]{0.6\textwidth}
\centering
\includegraphics[width=1.0\textwidth]{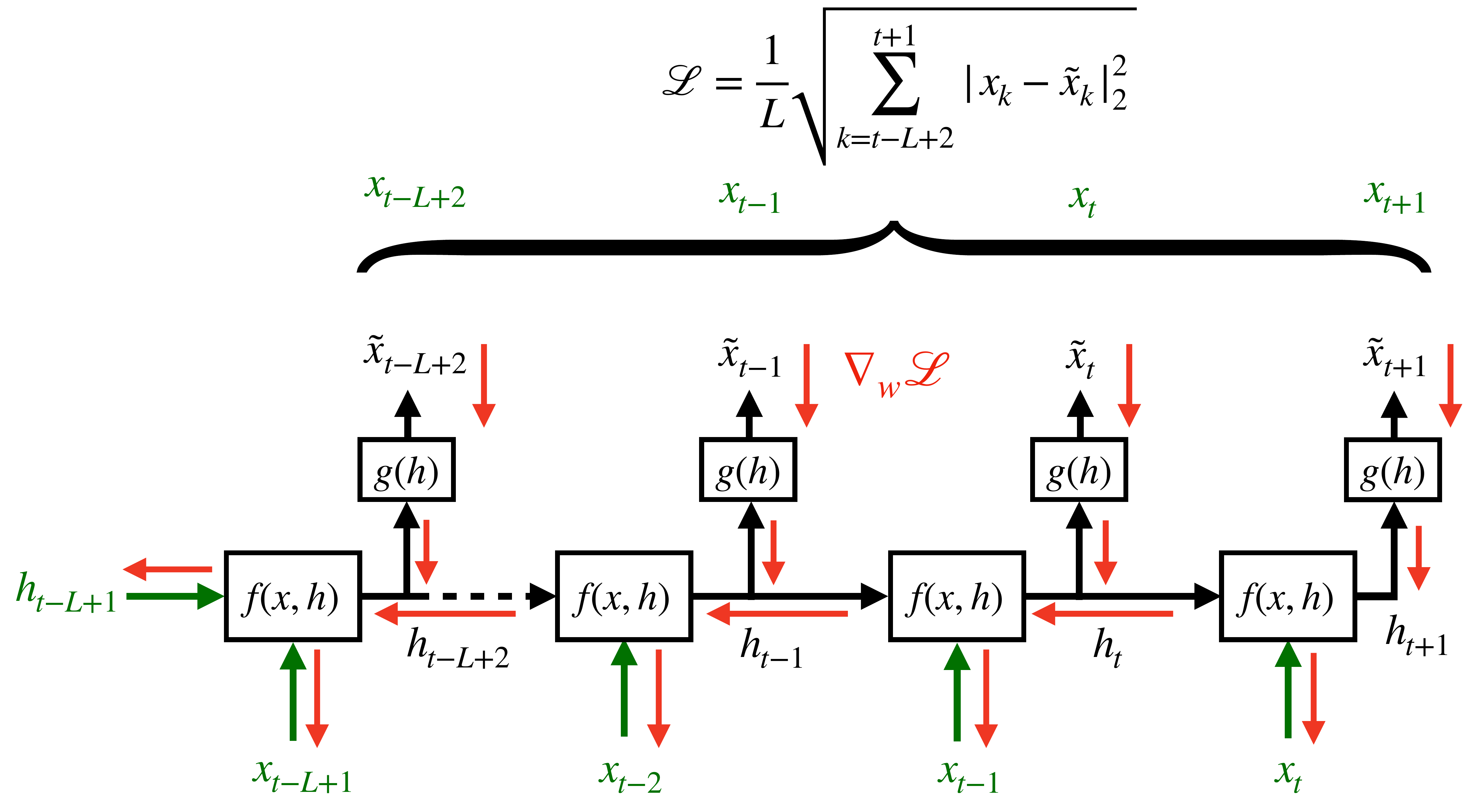}
\caption{BBTT-TF}
\label{fig:bptt:bptt-tf}
\hspace{0.3cm}
\end{subfigure}
\hfill 
\begin{subfigure}[tbhp]{0.6\textwidth}
\centering
\includegraphics[width=1.0\textwidth]{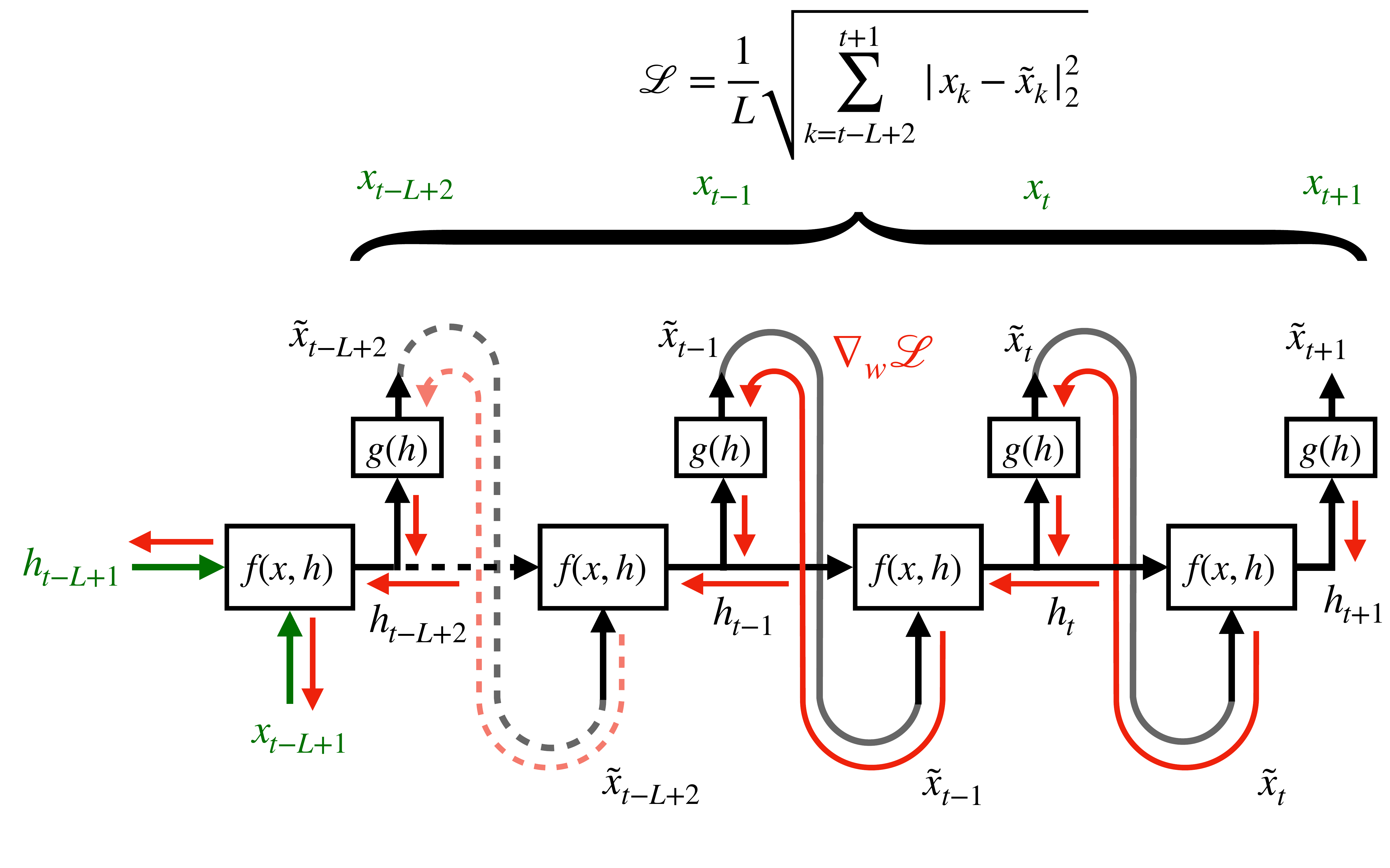}
\caption{BPTT-A}
\label{fig:bptt:bptt-a}
\hspace{0.3cm}
\end{subfigure}
\hfill 
\begin{subfigure}[tbhp]{0.6\textwidth}
\centering
\includegraphics[width=1.0\textwidth]{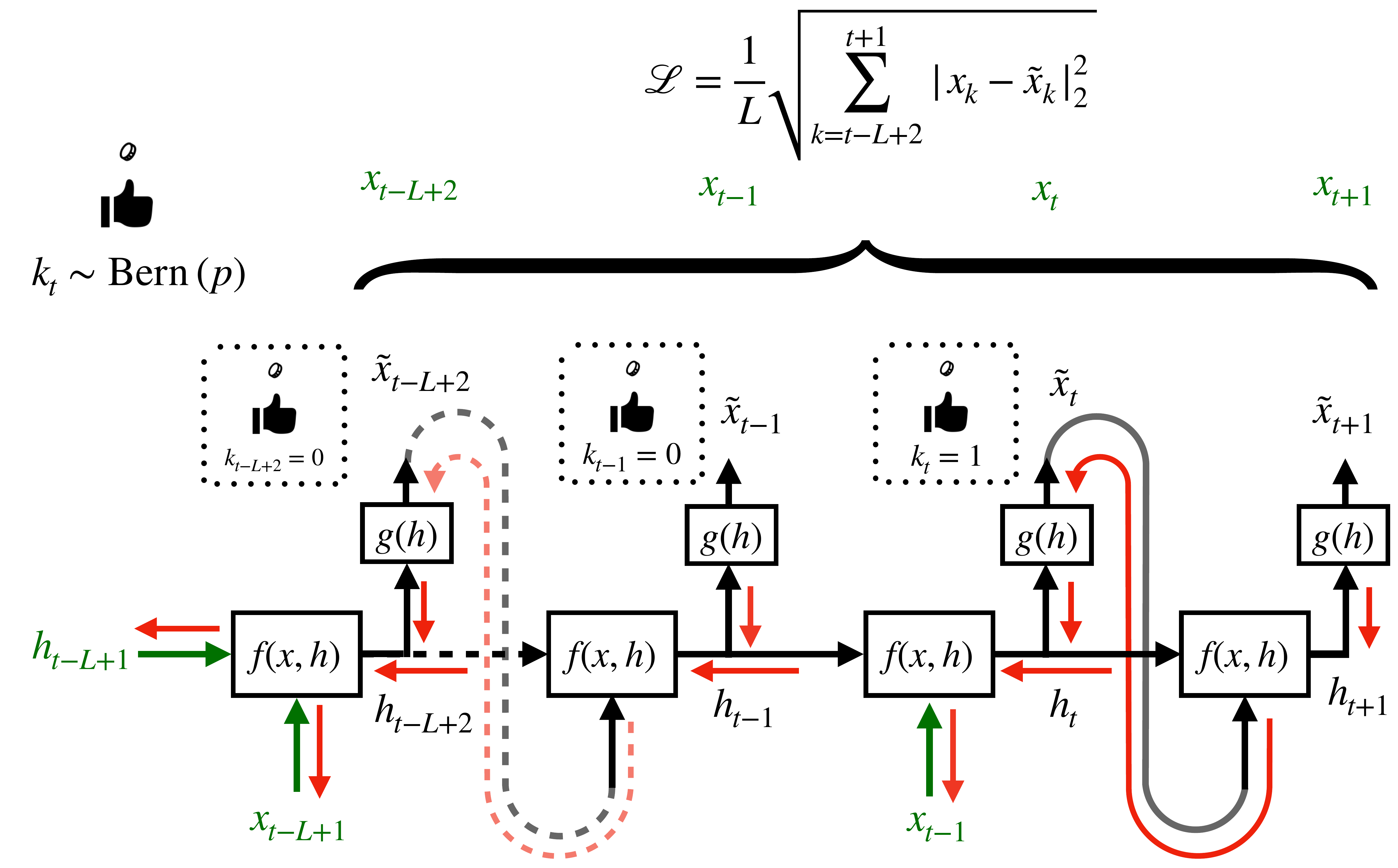}
\caption{BPTT-SA}
\label{fig:bptt:bptt-sa}
\hspace{0.3cm}
\end{subfigure}
\caption{
(a) Illustration of the forward pass and backward gradient flow of Backpropagation Through Time with Teacher Forcing (BPTT-TF), with green indicating the data (input and target). The forward pass of the network is represented by black arrows, while the backward gradient pass is depicted in red.
(b) In Autoregressive BPTT (BPTT-A) the networks' output is propagated through the network, only providing input data at the first timestep.
(c) In Scheduled Autoregressive BPTT (BPTT-SA) the behavior depends on the autoregressive probability $p$ parametrizing a Bernulli distribution.
}
\label{fig:bptt}
\end{figure}

In the "teacher forcing" approach we seek to make a prediction at $t+1$ given a stream of ground-truth data $\{ x_0, \dots, x_t, x_{t+1}\}$,  that is provided as input to the network. 
During the forward pass, the network is unrolled for $t+1$ timesteps, applying the~\cref{eq:rnn} iteratively. 
The training method is called Backpropagation Through Time with Teacher Forcing (BPTT-TF). 
In practice, due to memory and computational limitations, the input data stream for predicting the output at time $t+1$ is truncated to the last $L$ timesteps (sequence length), i.e., $\{ x_{t-L+1}, \dots, x_t\}$.
A schematic view of the computational graph of BPTT-TF and the backward flow of the gradient is shown in~\Cref{fig:bptt}.

The network's output $o_t$ (and thus the loss defined in~\cref{eq:loss}) is a function of the initialization of the hidden state $h_0$, the input stream, and the RNN weights $\{w_h, w_o \}$, i.e.,
\begin{equation}
o_t =
\operatorname{TF}
\Big(
\underbrace{
h_{0},
}_{\text{initial hidden state}}
\underbrace{
x_{0}
, \,
\dots  , \,
x_{t}
}_{\text{input stream}}
\,
\,
;
\,
\,
\underbrace{
w_h,
w_o
}_{\text{weights}}
\Big)
.
\label{eq:o:teacherforcing}
\end{equation}
The loss function is defined in~\cref{eq:loss}.
Following \cite{zhang2021dive}, the gradients of the loss with respect to the network's weights computed with backpropagation are given by:
\begin{gather}
\frac{
\partial \l(x_{t+1}, o_t)
}{
\partial w_o
}
= 
\frac{
\partial \l(x_{t+1}, o_t)
}{
\partial o_t
}
\frac{
\partial \g
}{
\partial w_o
},
\\
\frac{
\partial \l(x_{t+1}, o_t)
}{
\partial w_h
}
= 
\frac{
\partial \l(x_{t+1}, o_t)
}{
\partial o_t
}
\frac{
\partial \g
}{
\partial h_t
}
\hw
.
\label{eq:gradients}
\end{gather}
The term $\partial \l(x_{t+1}, o_t) / \partial o_t$ is evaluated by~\cref{eq:loss} and $\partial \g / \partial w_o$ by~\cref{eq:out}.
Evaluation of the term $\partial h_t / \partial w_h$ involves a  recurrence, as  $h_t$ depends on $h_{t-1}$ and $w_h$, while $h_{t-1}$ also depends on $w_h$.
The gradient can be evaluated using the chain rule
\begin{equation}
\hw
=
\fwh
+
\fh
\frac{ \partial h_{t-1} }{ w_h}
.
\label{eq:hw}
\end{equation}
For the evaluation of this gradient we utilize the following lemma, whose proof can be found in~\cref{app:sec:lemma}.
\begin{lemma}
Assume that there are three sequences $a_t, b_t$ and $c_t$, with $a_0=0$ and $a_t=b_t + c_t a_{t-1}$ for $t \in {1,2,\dots, T}$.
For $t>1$ it holds that:
\begin{equation}
a_t = b_t + \sum_{i=1}^{t-1}
\Big(
\prod_{j=i+1}^{t} c_j
\Big)
b_i.
\label{eq:lemma}
\end{equation}
\label{lemma:1}
\end{lemma}
By setting:
\begin{equation}
a_t = \hw, \quad
b_t = \fwh, \quad
c_t^{\operatorname{TF}} = c_t = \fh,
\label{eq:abc:tf}
\end{equation}
the three series satisfy the requirements of the lemma, namely $a_t=b_t + c_t^{\operatorname{TF}} a_{t-1}$ as per~\cref{eq:hw}, with $a_0=0$.
As a consequence, by substituting~\cref{eq:abc:tf} into~\cref{eq:lemma}, we end up with:
\begin{equation}
\hw
=
\fwh
+
\sum_{i=1}^{t-1}
\Bigg(
\prod_{j=i+1}^t
\frac{
\partial f(x_j, h_{j-1}; w_h)
}{
\partial h_{j-1}
}
\Bigg)
\frac{
\partial f(x_i, h_{i-1}; w_h)
}{
\partial w_h
}
.
\label{eq:gradient:tf}
\end{equation}
Note that the gradient in~\cref{eq:gradient:tf} is relevant when the input sequence in the forward pass consists of ground-truth data.
A common problem encountered in practice during RNN training is vanishing and exploding gradients during backpropagation~\citep{le2016quantifying,hochreiter1998vanishing}.
The gradient~\cref{eq:gradient:tf} entails the product of $c_t^{\operatorname{TF}}$s (defined in~\cref{eq:abc:tf}).
Successful training with BPTT (i.e., capturing long-term dependencies, loss reduction, informative gradients, non-oscillatory loss behavior) depends on keeping this gradient at a reasonable norm.

In the truncated BPTT-TF the gradient~\cref{eq:gradient:tf} is not computed over the whole sequence.
The forward propagation is performed over a subset of the input stream consisting of the last $L$ steps, and the backpropagation is truncated ignoring the history prior to these $L$ timesteps.
The hidden state $h_0=h_{t-L+1}$ at the point where the gradient flow is truncated can be set from the previous batch (state-full RNN), or set to zero in state-less RNNs.
Here we consider the state-full RNN case.

\subsection{Backpropagation Through Time with Autoregression}
\label{sec:sub:auto}

In the previous section (\cref{sec:sub:tf}), it was explained that when a network is trained with teacher forcing, it learns to predict the state evolution at a future timestep for a specific lead time. 
However, in practical applications, it is often necessary for the network to provide forecasts for multiple lead times or to forecast the future evolution of a data stream.
Training multiple models for different lead times is computationally expensive and not scalable, and it is preferable to use a trained model for future data stream trajectory forecasts. 
Autoregressive inference can achieve this by feeding the output of the RNN back into the input, iteratively forecasting the data stream's evolution. 
However, using teacher forcing in this case poses a significant problem, as the network was not trained to predict its own outputs at the input, rather than ground-truth data. 
Additionally, the network outputs might follow a different distribution than the ground-truth data due to imperfect training. 

We address these challenges by proposing an alternative training method called Autoregressive BPTT and derive the associated gradient.
In Autoregressive BPTT, in order to  compute the output at timestep $t$, the network iteratively propagates its own predictions, beginning with an initial hidden state $h_0$ and the first input data state of the stream $x_0$. 
The output of the network is determined by the weights of the network and its initial state, $h_0$ and $x_0$, i.e.,
\begin{gather}
o_{t}
=
\operatorname{IF} \big(
\underbrace{
h_{0},
}_{\text{initial state}}
\underbrace{
x_{0}
}_{\text{initial input}}
\,
\,
;
\,
\,
\underbrace{
w_h,
w_o
}_{\text{weights}}
\big)
.
\label{eq:o:auto}
\end{gather}
Note the difference compared to the equation of the output in teacher forcing in~\cref{eq:o:teacherforcing}.
The gradient of the loss with respect to the network's parameters in this case is different.
We start by repeating the gradients from~\cref{eq:gradients}:
\begin{gather*}
\frac{
\partial \l(x_{t+1}, o_t)
}{
\partial w_o
}
= 
\frac{
\partial \l(x_{t+1}, o_t)
}{
\partial o_t
}
\frac{
\partial \g
}{
\partial w_o
}
\\
\frac{
\partial \l(x_{t+1}, o_t)
}{
\partial w_h
}
= 
\frac{
\partial \l(x_{t+1}, o_t)
}{
\partial o_t
}
\frac{
\partial \g
}{
\partial h_t
}
\hw
\end{gather*}
In the following, we treat the input $x_t$ as a function of the previous hidden state $x_t=g(h_{t-1})$ due to autoregression.
This implies:
\begin{equation}
\begin{split}
\hw
&=
\fwh
+
\fx
\underbrace{
\xh
\hmw
}
_{
\xw
}
+
\fh
\hmw 
\implies \\
&=
\fwh
+
\Bigg(
\fx
\xh
+
\fh
\Bigg)
\hmw
\end{split}
\label{eq:auto:hw}
\end{equation}
To evaluate the recurrence relation and evaluate the gradient in~\cref{eq:auto:hw}, we use again lemma~\ref{lemma:1}.
By setting
\begin{equation}
a_t = \hw, \quad
b_t = \fwh, \quad
c_t^{\operatorname{AR}} = c_t = \fx
\xh
+
\fh
\label{eq:abc:ar}
\end{equation}
the three series satisfy the requirements of the lemma, namely $a_t=b_t + c_t^{\operatorname{AR}} a_{t-1}$ as per~\cref{eq:auto:hw}, with $a_0=0$.
As a consequence, by substituting~\cref{eq:abc:ar} into~\cref{eq:lemma}, we end up with:
\begin{equation}
\hw
=
\fwh
+
\sum_{i=1}^{t-1}
\Bigg(
\prod_{j=i+1}^t
\frac{
\partial f(x_j, h_{j-1}; w_h)
}{
\partial x_{j}
}
\frac{
\partial g(h_{j-1})
}{
\partial h_{j-1}
}
+
\frac{
\partial f(x_j, h_{j-1}; w_h)
}{
\partial h_{j-1}
}
\Bigg)
\frac{
\partial f(x_i, h_{i-1}; w_h)
}{
\partial w_h
}
\label{eq:gradient:ar}
\end{equation}
In contrast to the gradient in the teacher forcing case 
(~\cref{eq:gradient:tf}), the gradient in the autoregressive case contains a product that also involves the hidden-to-output mapping, i.e. $\partial g(h_{j-1}) / \partial h_{j-1}$.
As a consequence, assuming no vanishing or exploding gradients, the hidden-to-output mapping is also regularized.
Previous studies on forecasting high dimensional dynamical systems, have reported that RNNs trained with teacher forcing tend to produce unrealistic patterns and diverge from the underlying attractors in autoregressive testing~\citep{vlachas2020learning}.
We argue that one of the main reasons of this degeneracy is that the teacher forcing is not regularizing the hidden-to-output mapping, as it is not involved in the product (recursive time unrolling) in the gradient.
As a consequence, although predictions are accurate on the short-term, the network's output weights are not regularized for the iterative propagation of the output, causing divergence of the predictions.
This effect is more prominent in long-term spatiotemporal forecasting of high-dimensional dynamical systems, where the hidden-to-output mapping can be a large Convolutional Neural Network (CNN) with thousands of parameters.
A schematic view of Autoregressive BPTT (BPTT-A) is given in~\Cref{fig:bptt:bptt-a}.

\subsection{Truncated Backpropagation Through Time with Scheduled Autoregression}
\label{sec:sub:sa}

It is worth noting that prior to training an RNN, its weights are initialized randomly and the untrained network is not capable of generating accurate short-term predictions. 
Propagating these imprecise predictions in an iterative manner would result in a gradual buildup of errors. 
For this reason, utilizing the autoregressive gradient during the initial training period is not a justifiable approach since the network is yet to be trained and is inaccurate, even on short-term predictions.
For this reason, we propose a scheduled autoregressive approach, inspired by~\cite{bengio2015scheduled}.

In Scheduled Autoregressive BPTT (BPTT-SA) the selection of the propagation type (autoregressive or teacher forcing) at each timestep depends on a sample $k_t \in \{0,1\}$ from a Bernulli distribution, parametrized by the iterative forecasting probability $p$, i.e. $k_t \sim \operatorname{Bern}(p)$.
$k_t$ is one with probability $p$ and zero with probability $1-p$ (coin-flip).
At each time-step $t$ we sample a different $k_t$ and decide on BPTT-TF if $k_t=0$ or BPTT-A if $k_t=1$.
Following the argumentation in~\cref{sec:sub:tf} and ~\cref{sec:sub:auto}, the gradient $\hw$ is evaluated as:
\begin{equation}
\hw
=
\fwh
+
\sum_{i=1}^{t-1}
\Bigg(
\prod_{j=i+1}^t
\big(
c_j^{\operatorname{TF}}
\big)^{(1-k_j)}
\,
\big(
c_j^{\operatorname{AR}}
\big)^{k_j}
\Bigg)
\frac{
\partial f(x_i, h_{i-1}; w_h)
}{
\partial w_h
}
.
\label{eq:gradient:sa}
\end{equation}
Note that for $p=0$, $k_j\sim \operatorname{Bern}(0)=1$, the product $
\big(
c_j^{\operatorname{TF}}
\big)^{(1-k_j)}
\,
\big(
c_j^{\operatorname{AR}}
\big)^{k_j}$ in~\cref{eq:gradient:sa} is equal to $c_j^{\operatorname{TF}}$ and the gradient in~\cref{eq:gradient:sa} evaluates to the BPTT-TF gradient expressed in~\cref{eq:gradient:tf}.
In contrast, for $p=1$, the aforementioned product is equal to $c_j^{\operatorname{AR}}$ and the gradient evaluates to the BPTT-A gradient expressed in~\cref{eq:gradient:ar}.

\begin{figure}[pos=H]
\centering
\includegraphics[width=0.6\textwidth]{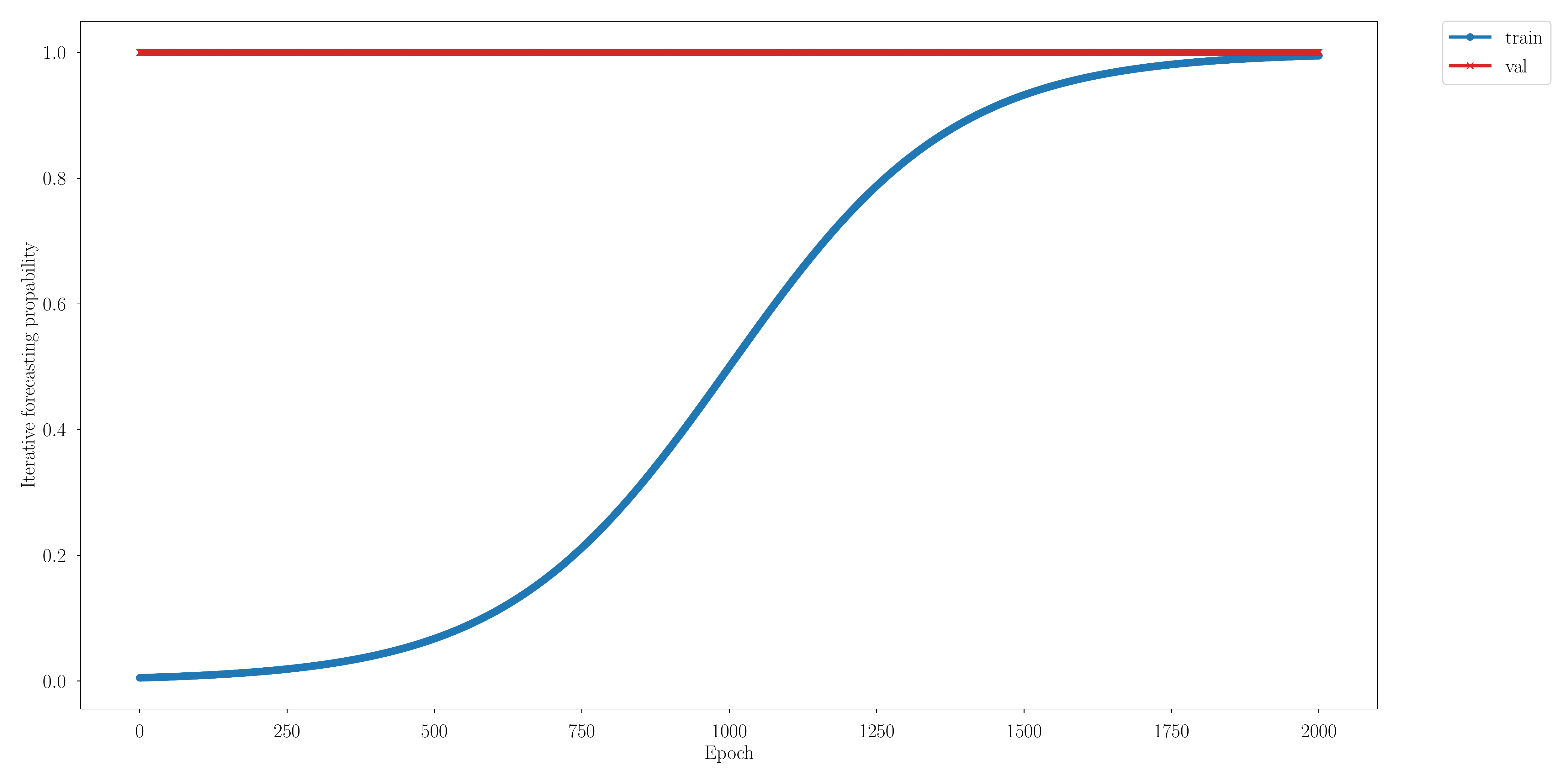}
\captionof{figure}{Schedule of iterative forecasting probability $p$.}
\label{fig:schedule_loss}
\end{figure}

During training, $p$ follows an inverse sigmoid schedule for the train loss.
At initial training epochs $p$ is close to zero, and the model is trained with the standard BPTT loss (equivalent to BPTT-TF).
As training progresses, $p$ is gradually annealed till it reaches $p \to 1$ towards the final training epochs, leading to BPTT-A.
The validation loss is computed for $p=1$ (equivalent to BPTT-A).
This ensures that the validation loss according to which we pick the optimal model is the autoregressive loss.
The schedule is depicted in~\Cref{fig:schedule_loss} for a training procedure of $2000$ epochs in total.

\section{Results}

We benchmark the proposed BPTT-SA with standard BPTT-TF and the method proposed in~\cite{bengio2015scheduled} denoted as BPTT-SS (scheduled sampling).
All models are implemented in Pytorch~\citep{paszke2019pytorch}, and ported to a single Nvidia Tesla P100 GPU.
In all experiments, the models are trained with the Adam~\citep{kingma2014adam} optimizer, and we employ validation based early stopping to cope with overfitting.

\subsection{Mackey-Glass}
\label{sec:mackey}

We evaluate the effectiveness of BPTT-SA in forecasting chaotic time series from the Mackey-Glass (MG) equations.
This is a challenging benchmark problem due to its chaotic nature~\citep{voelker2019legendre,gers2002applying}.
The time-series is generated by the delay differential equation
\begin{equation}
\frac{dx}{dt} = \frac{\alpha \, x(t- \tau) }{1 + x^n(t-\tau)} - \beta x(t).
\label{eq:mg}
\end{equation}
We consider the parameter setting $\alpha=0.2$, $\beta=0.1$, $c=10$, and $\tau=17$.

The maximum Lyapunov exponent, calculated with the method of~\cite{vlachas2020backpropagation} is $\Lambda_1 \approx 8.9 \cdot 10^{-3}$, leading to a Lyapunov time of $T^{\Lambda_1}=112$ time units.
We integrate~\Cref{eq:mg} with a fourth order Runge-Kutta scheme with $\delta t=0.1$ up to $T=2\cdot 10^5$.
The data are subsampled after integration to $\Delta t =1.0$.
$32$ sequences of $1120$ timesteps each ($10$ Lyapunov times) are generated for training.
A data set of the same size is considered for validation.
The remaining data are considered for testing.
In order to test the proposed algorithm in the autoregressive setting, $100$ initial conditions are randomly sampled from the test-data and the networks are asked to forecast the next $896$ steps, that amounts to approximately $8$ Lyapunov times (after an initial warm-up period of 20 timesteps).
As comparison metrics, we consider the Root Mean Square Error (RMSE) and the error on the power spectrum (frequency content).
Moreover, we consider two different noise levels on the data, a signal-to-noise ratio of $\operatorname{SNR}=60$, and a disturbed case of $\operatorname{SNR}=10$.

\begin{figure}[pos=H]
\centering
\begin{subfigure}[tbhp]{0.24\textwidth}
\centering
\includegraphics[width=1.0\textwidth,clip]{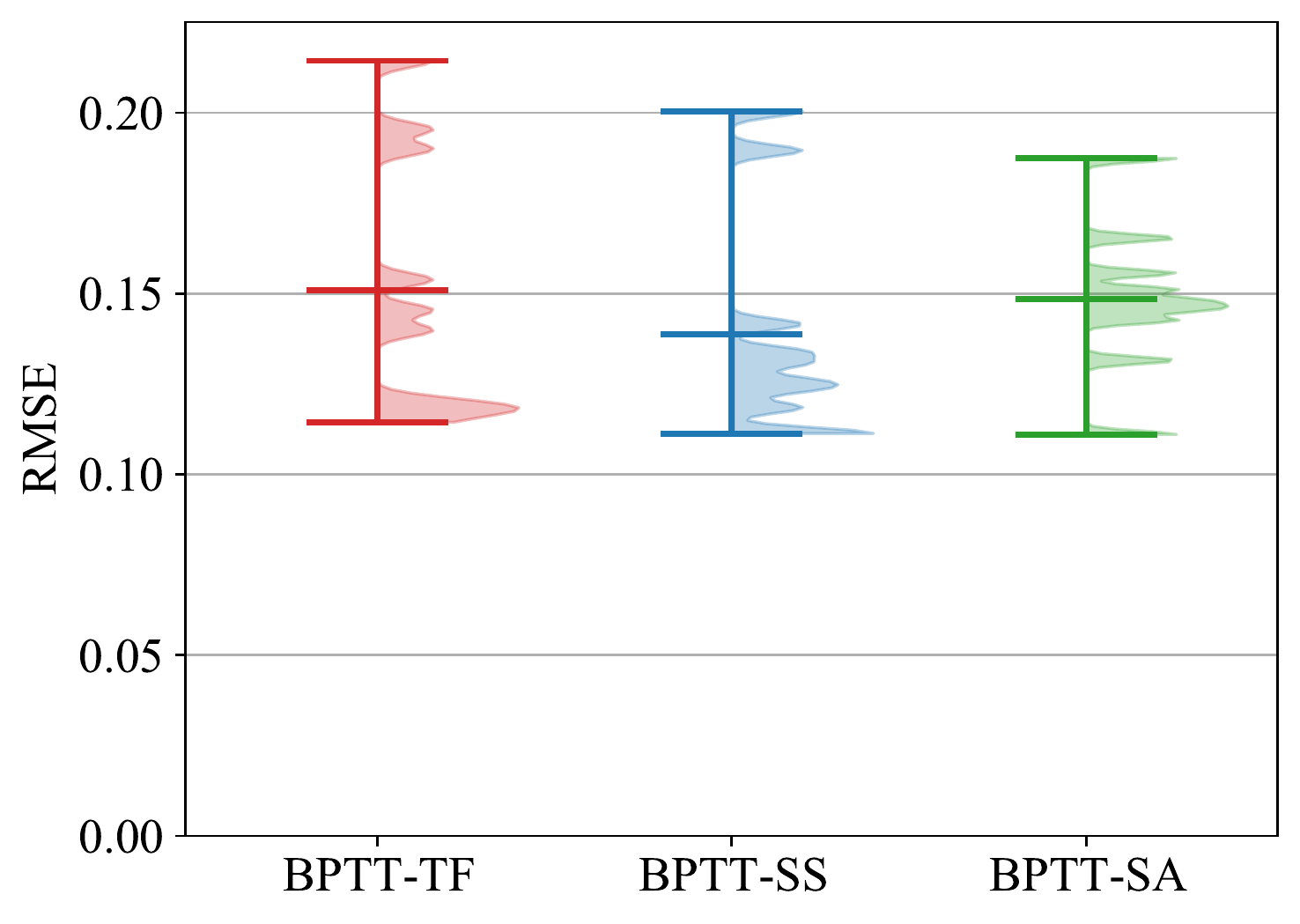}
\caption{$\operatorname{SNR}=60$}
\label{fig:MackeyGlassSNR60:RMSE_ISF}
\end{subfigure}
\hfill 
\begin{subfigure}[tbhp]{0.24\textwidth}
\centering
\includegraphics[width=1.0\textwidth,clip]{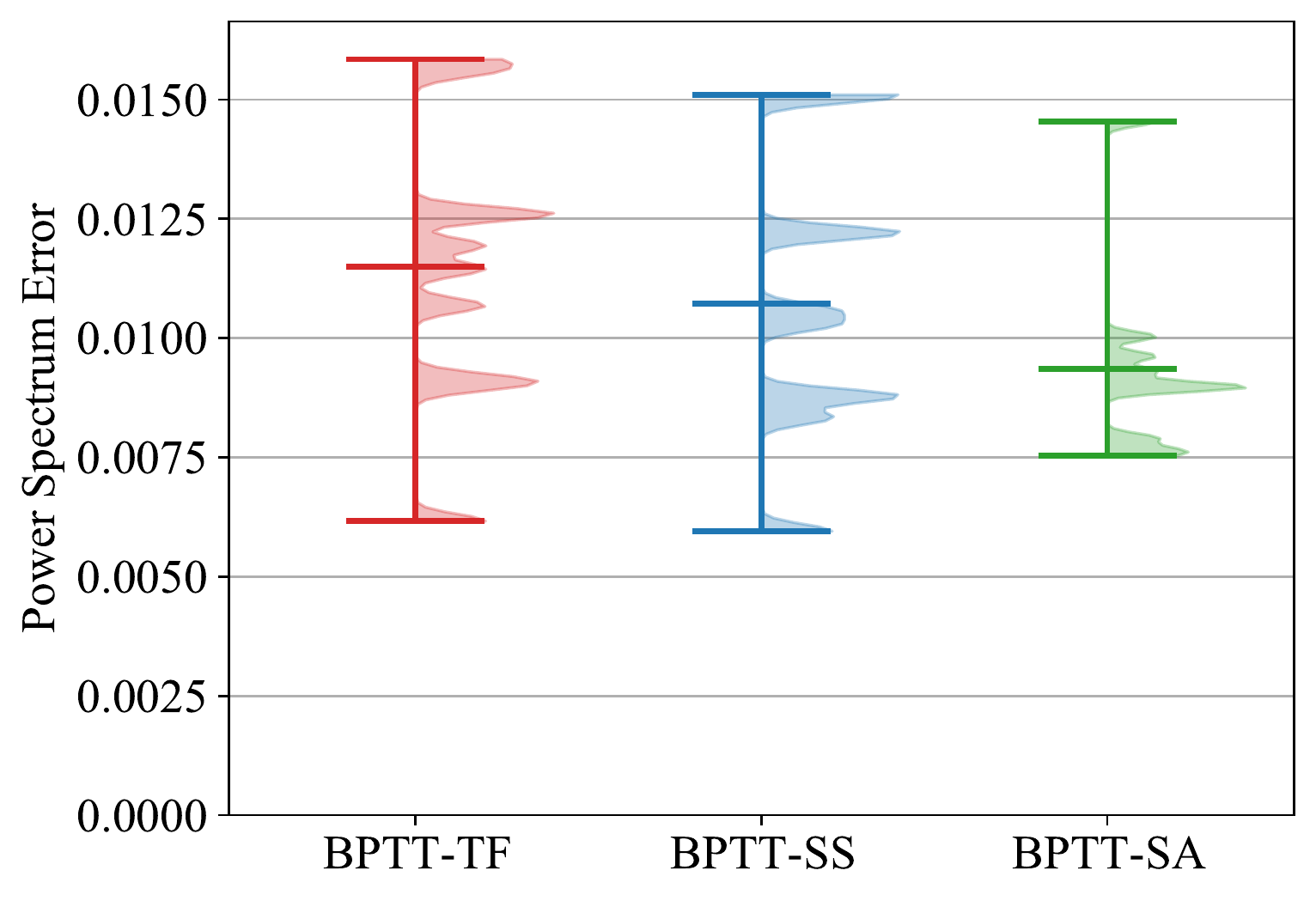}
\caption{$\operatorname{SNR}=60$}
\label{fig:MackeyGlassSNR60:FREQ_ISF}
\end{subfigure}
\hfill 
\begin{subfigure}[tbhp]{0.24\textwidth}
\centering
\includegraphics[width=1.0\textwidth,clip]{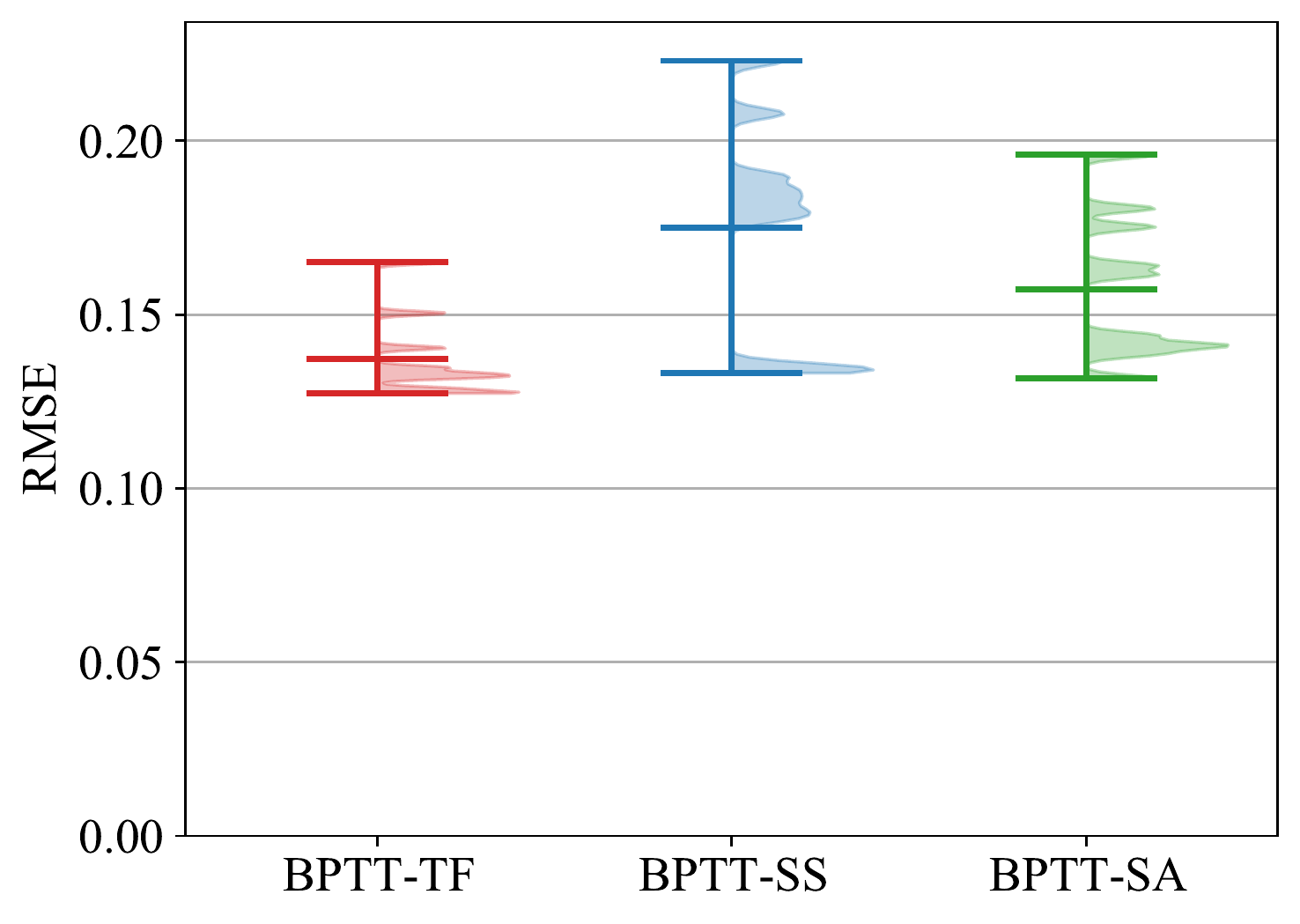}
\caption{$\operatorname{SNR}=10$}
\label{fig:MackeyGlassSNR10:RMSE_ISF}
\end{subfigure}
\hfill 
\begin{subfigure}[tbhp]{0.24\textwidth}
\centering
\includegraphics[width=1.0\textwidth,clip]{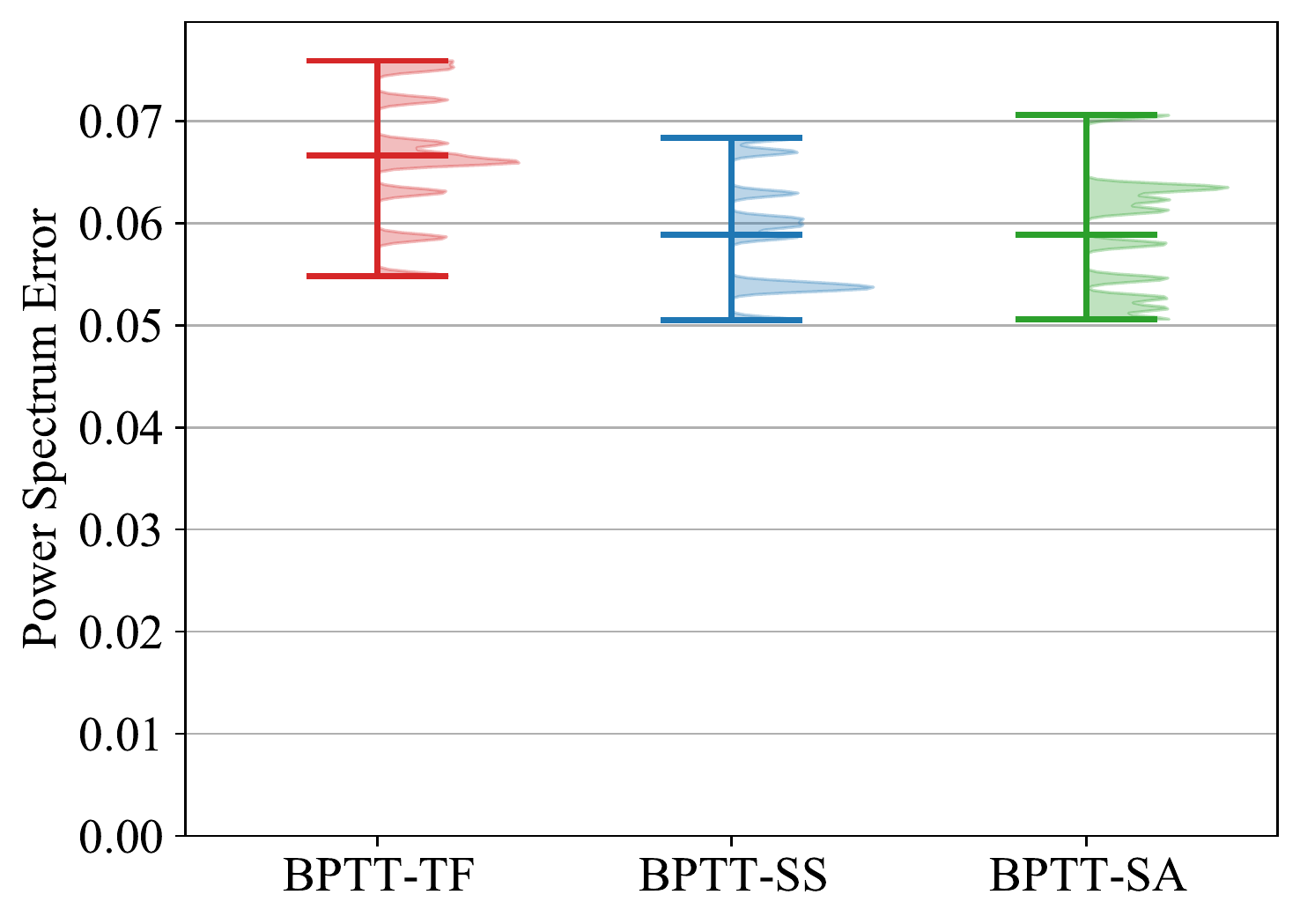}
\caption{$\operatorname{SNR}=10$}
\label{fig:MackeyGlassSNR10:FREQ_ISF}
\end{subfigure}
\centering
\caption{
Evaluation of the performance of the training methods in forecasting the long-term dynamics of the Mackey-Glass time-series.
A prediction horizon of $896$ time steps is considered, and results are averaged over $100$ initial conditions in the test data.
}
\label{fig:results:mackeyglass}
\end{figure}

The results are illustrated in~\Cref{fig:results:mackeyglass}.
In the case of low level noise ($\operatorname{SNR}=60$), all methods show approximately the same performance in terms of the RMSE.
BPTT-SA shows slightly better performance on average and smaller variations between the different seeds (increased robustness) on the power spectrum error.
However, the differences are small.
In the $\operatorname{SNR}=10$ noise level, all three methods exhibit similar errors.

\subsection{Darwin Sea Level Pressure Dataset}
\label{app:sec:darwin}

\begin{figure}[pos=H]
\centering
\includegraphics[width=0.45\textwidth]{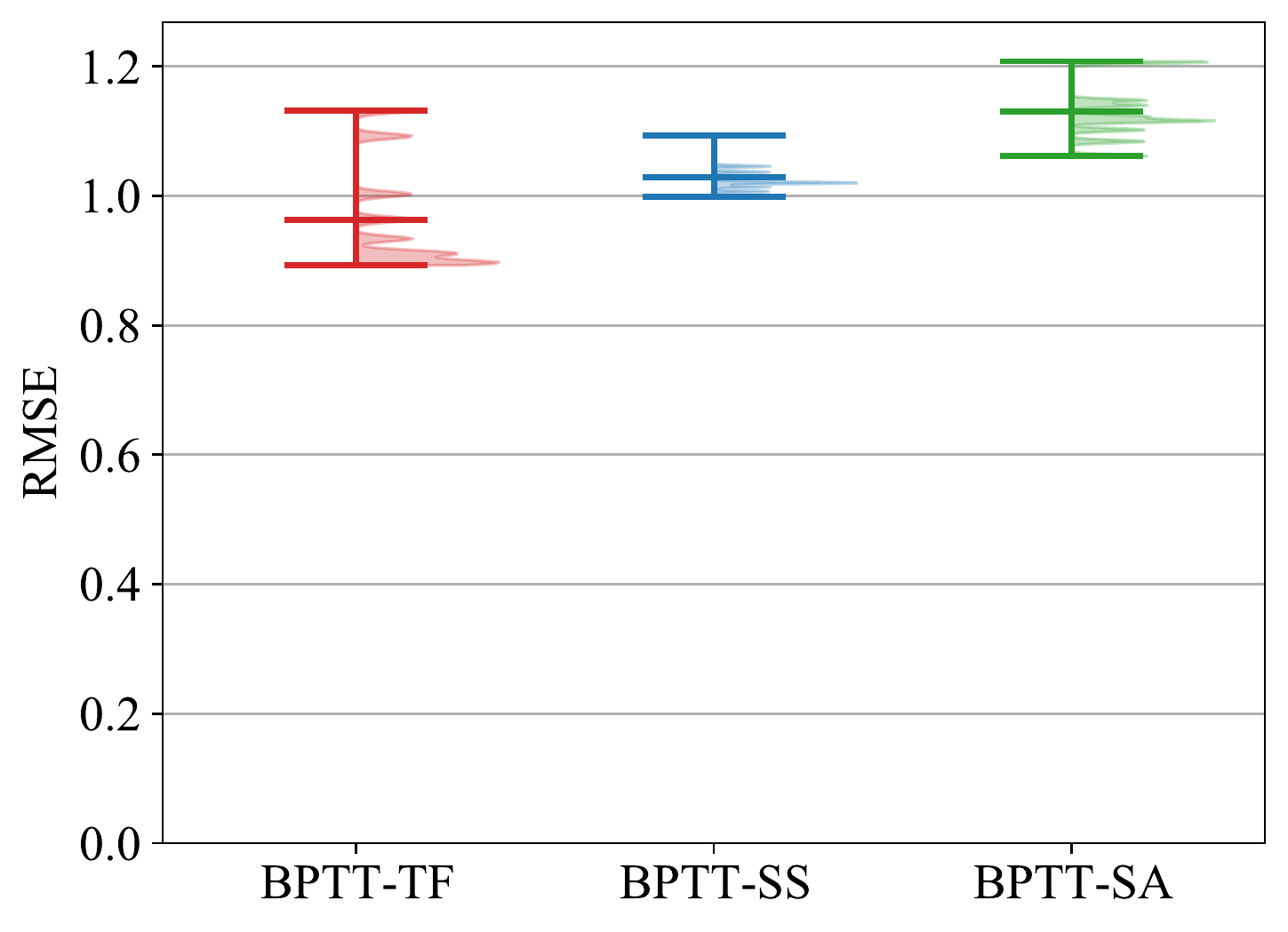}
\hfill
\includegraphics[width=0.45\textwidth]{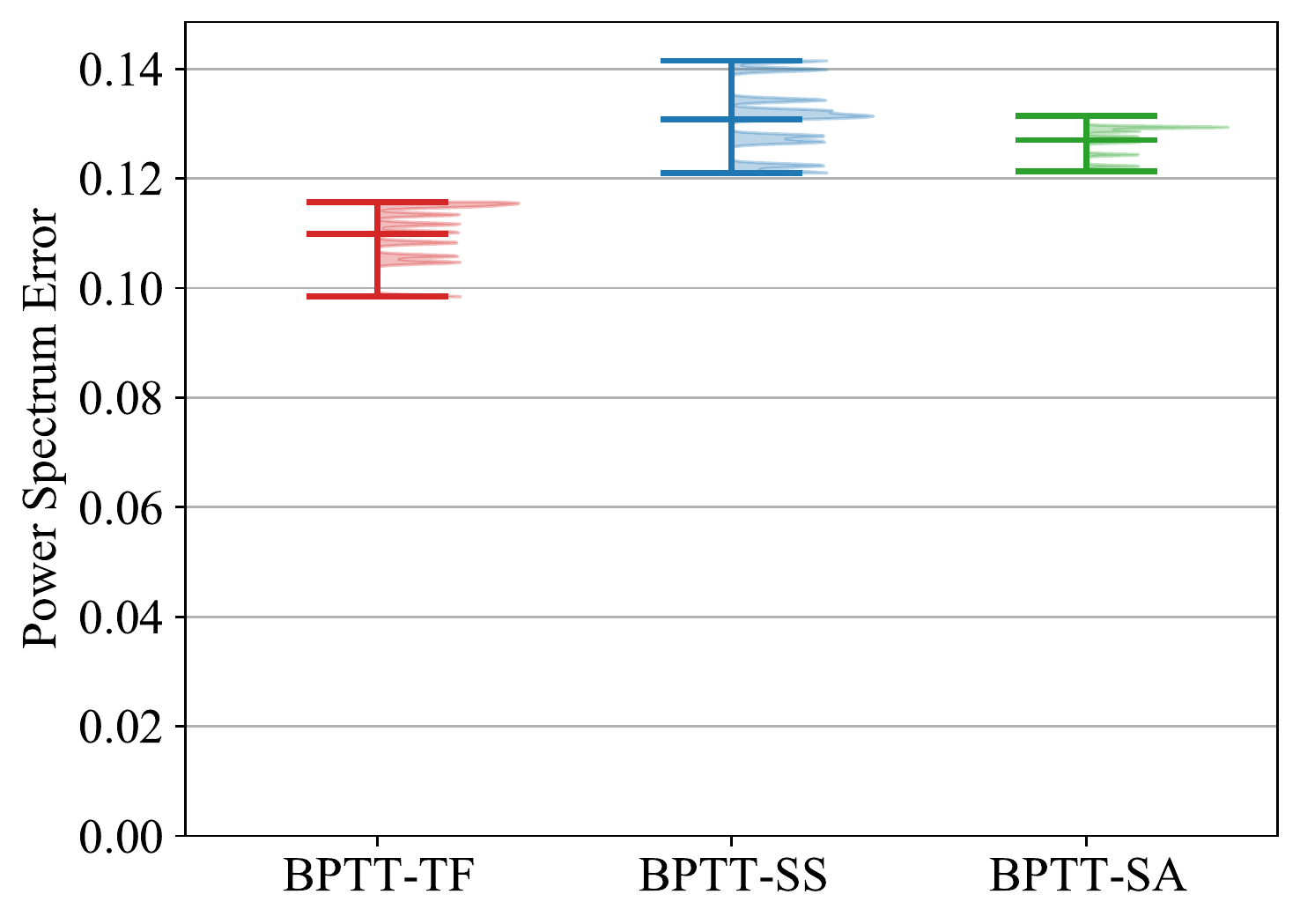}
\captionof{figure}{Results on the Darwin sea level pressure dataset.}
\label{fig:darwin:results}
\end{figure}

In this study, we conduct an evaluation of the long-term predictability of the methods using an open-source dataset that is widely utilized as a benchmark for time-series prediction techniques. 
Specifically, we examine the monthly average sea level pressure data recorded at Darwin between 1882 and 1998~\citep{harrison1997darwin}.
The dataset consists of 1400 samples.
The first 600 samples are used for training, and the next 400 for validation.
The long-term forecasting accuracy of the methods is evaluated on 32 initial conditions randomly sampled from the test data.
The RNNs forecast up to a prediction horizon of 100 timesteps, after an initial warm-up period of 50 timesteps. 
The results are illustrated in~\Cref{fig:darwin:results}.
We observe that both variants BPTT-SA and BPTT-SS do not offer any improvements in RMSE or in the power spectrum error.

\subsection{Navier-Stokes Equations: Flow past a circular cylinder at Re=200}
\label{sec:ns}

We consider long-term forecasting of the dynamics of the incompressible Navier-Stokes equations governing the flow past a circular cylinder in a channel in Reynolds number $Re=200$ in two dimensions.
The flow exhibits a periodic vortex street, so long-term prediction of the motion is possible.
Moreover, the dimensionality of the intrinsic dynamics (effective degrees of freedom) of the motion is low, as the motion can be characterized by a few dominant modes.
The latter implies that snapshots of the flow can be mapped to a reduced order latent space, representing the manifold of the effective dynamics~\citep{vlachas2022multiscale}.

We evaluate the effectiveness of BPTT-SA in two types of recurrent neural networks, Convolutional RNNs (ConvRNN) and Convolutional Autoencoder RNNs (CNN-RNNs).
The latter first identify a latent reduced order representation encoding the intrinsic dimensionality of the fluid flow, and learn the temporal dynamics on the latent space.
In contrast, ConvRNNs are replacing the operations on the RNN cell with convolutions while keeping the gating mechanisms~\citep{shi2015convolutional, kumar2020convcast} and do not require low dimensional intrinsic dynamics.

The flow is simulated with a Finite Element (FEM) solver~\citep{alnaes2015fenics} with a time-step of $0.001$.
The velocity $u \in \mathbb{R}^2$ and pressure $p \in \mathbb{R}$ values were extracted in a uniform grid of $160 \times 32$, and data are subsampled to $\Delta t = 0.01$.
For plotting purposes, we plot the Frobenius norm of the velocity $u=\sqrt{u_x^2 + u_y^2}$.
For more information on the geometry and simulation details refer to~\cite{petter2017solving}.
The total simulation time is $T=24$.
The first $200$ timesteps are used for training, the next $200$ for validation and the next $2000$ for testing.
The long-term iterative prediction performance of the methods is evaluated on prediction of $1000$ timesteps starting from $10$ initial conditions randomly sampled from the test data.
For more information about the hyperparameter tuning refer to~\Cref{app:ns:hyp}.

\begin{figure}[pos=H]
\centering
\begin{subfigure}[tbhp]{0.45\textwidth}
\centering
\includegraphics[width=1.0\textwidth]{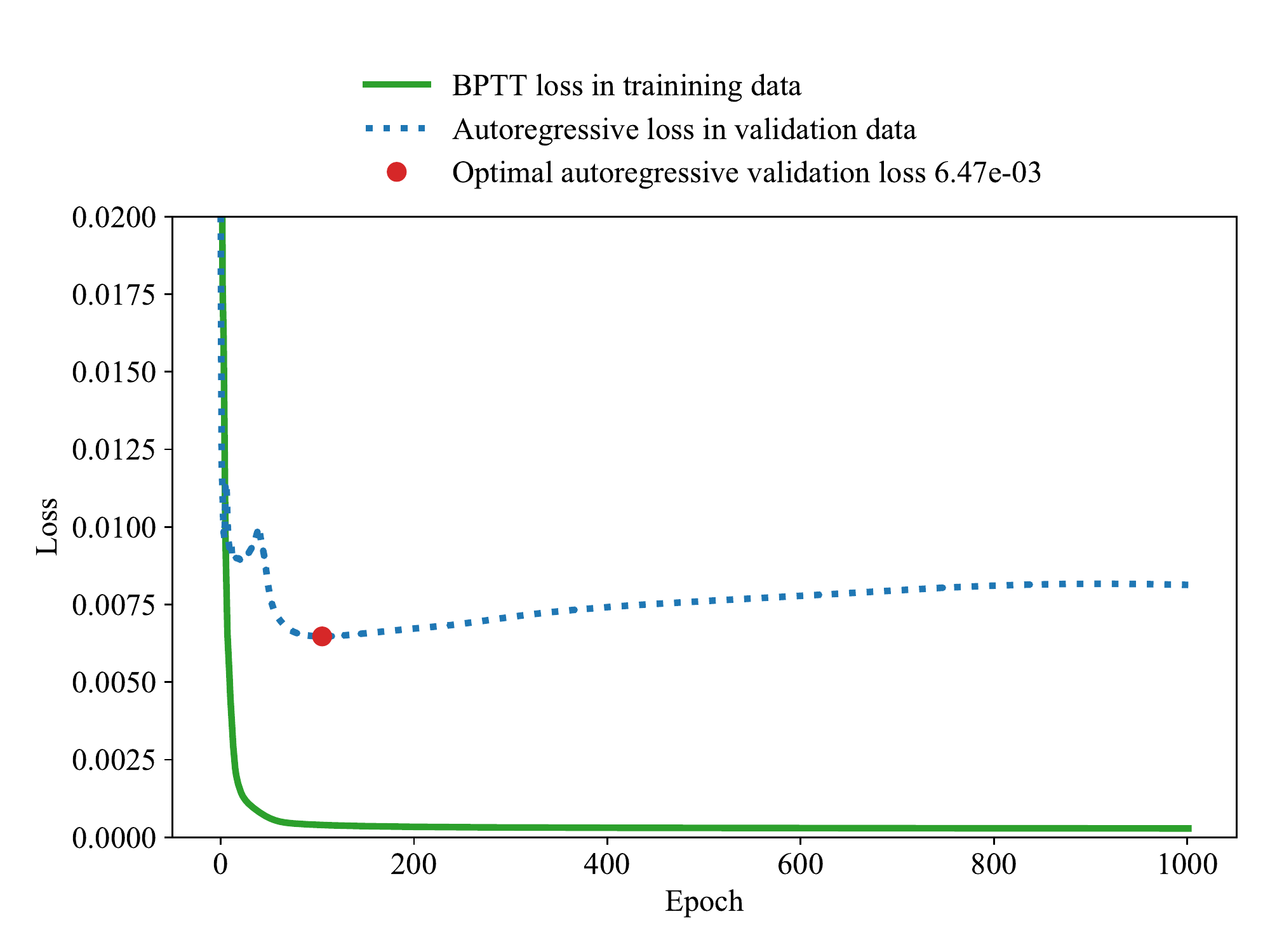}
\caption{CNN-RNN trained with BPTT}
\label{fig:NSC600_CNNRNN:Comparison_losses_BPTT}
\end{subfigure}
\\
\begin{subfigure}[tbhp]{0.45\textwidth}
\centering
\includegraphics[width=1.0\textwidth]{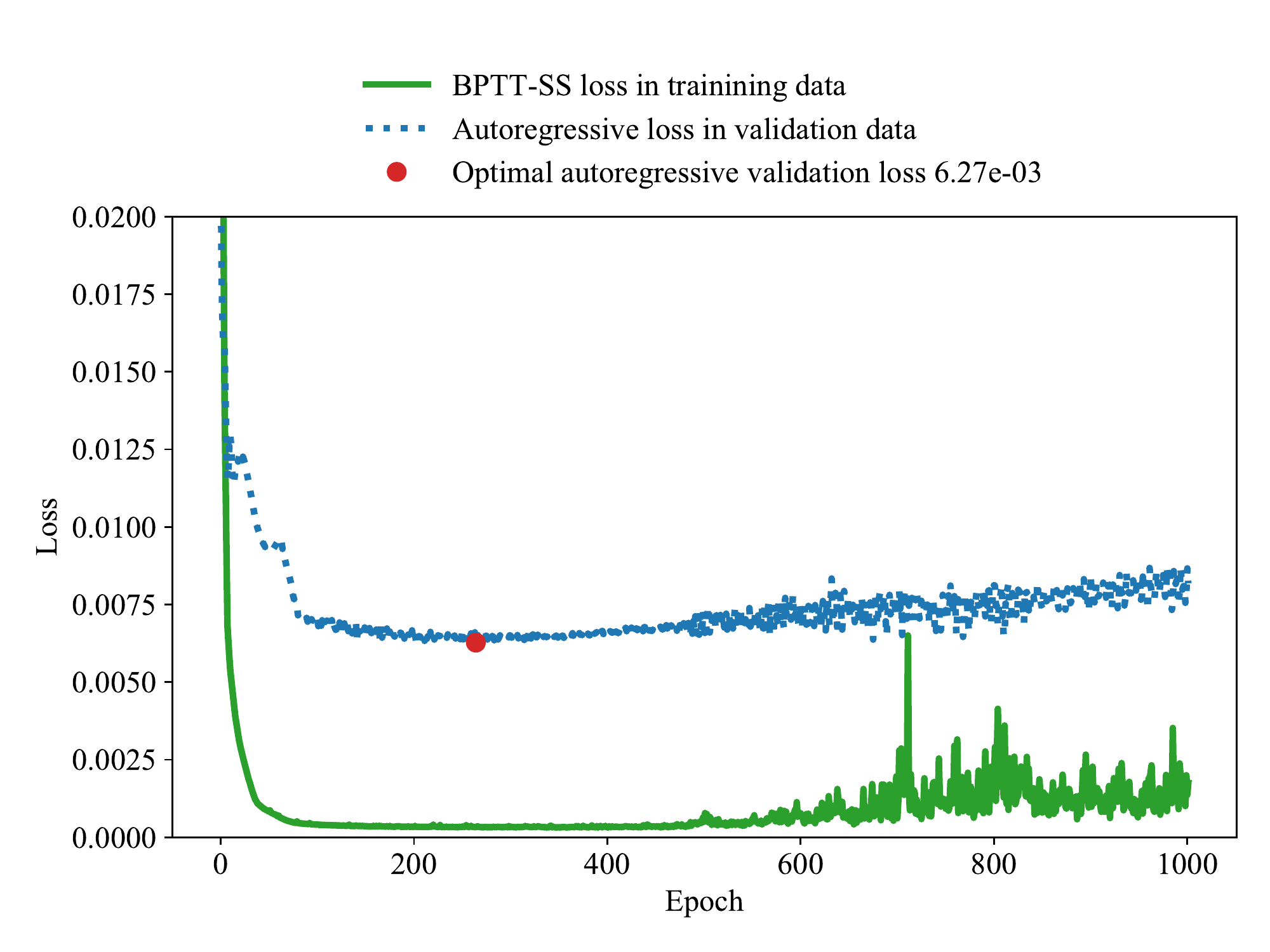}
\caption{CNN-RNN trained with BPTT-SS}
\label{fig:NSC600_CNNRNN:Comparison_losses_BPTT-SS}
\end{subfigure} 
\\
\begin{subfigure}[tbhp]{0.45\textwidth}
\centering
\includegraphics[width=1.0\textwidth]{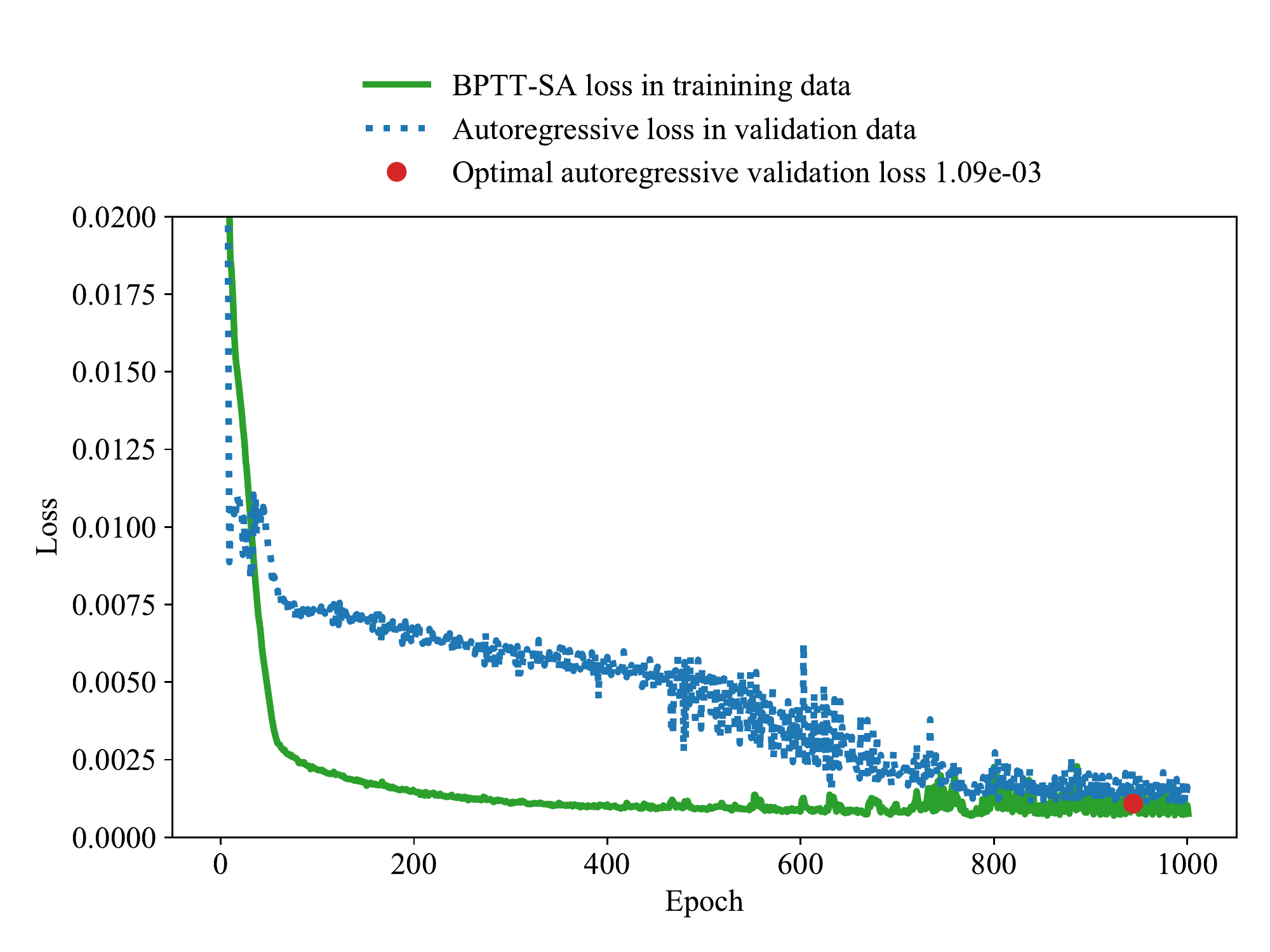}
\caption{CNN-RNN trained with BPTT-SA}
\label{fig:NSC600_CNNRNN:Comparison_losses_BPTT-SA}
\end{subfigure}
\caption{
Evolution of the training and validation losses in CNN-RNN training in the Navier-Stokes dataset.
}
\label{fig:results:nsc:training:cnnrnn}
\end{figure}

\begin{figure}[pos=H]
\centering
\begin{subfigure}[tbhp]{0.75\textwidth}
\centering
\includegraphics[width=1.0\textwidth]{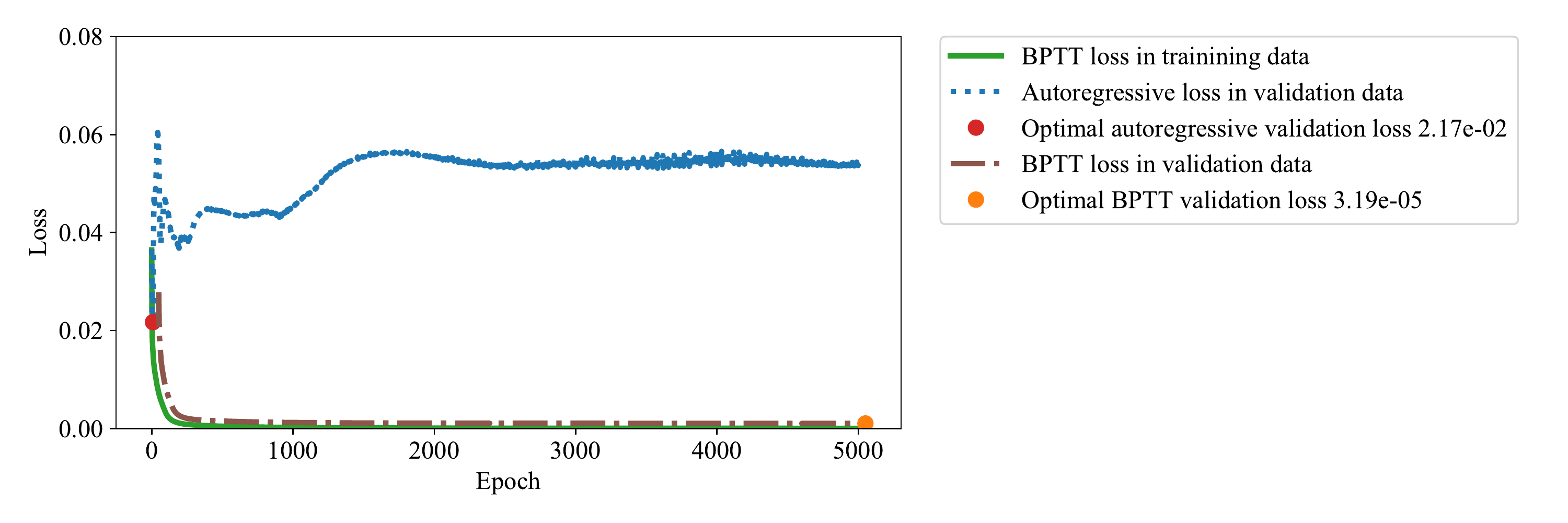}
\caption{ConvRNN trained with BPTT}
\label{fig:NSC600_ConvRNN:Comparison_losses_BPTT}
\end{subfigure}
\hfill 
\begin{subfigure}[tbhp]{0.75\textwidth}
\centering
\includegraphics[width=1.0\textwidth]{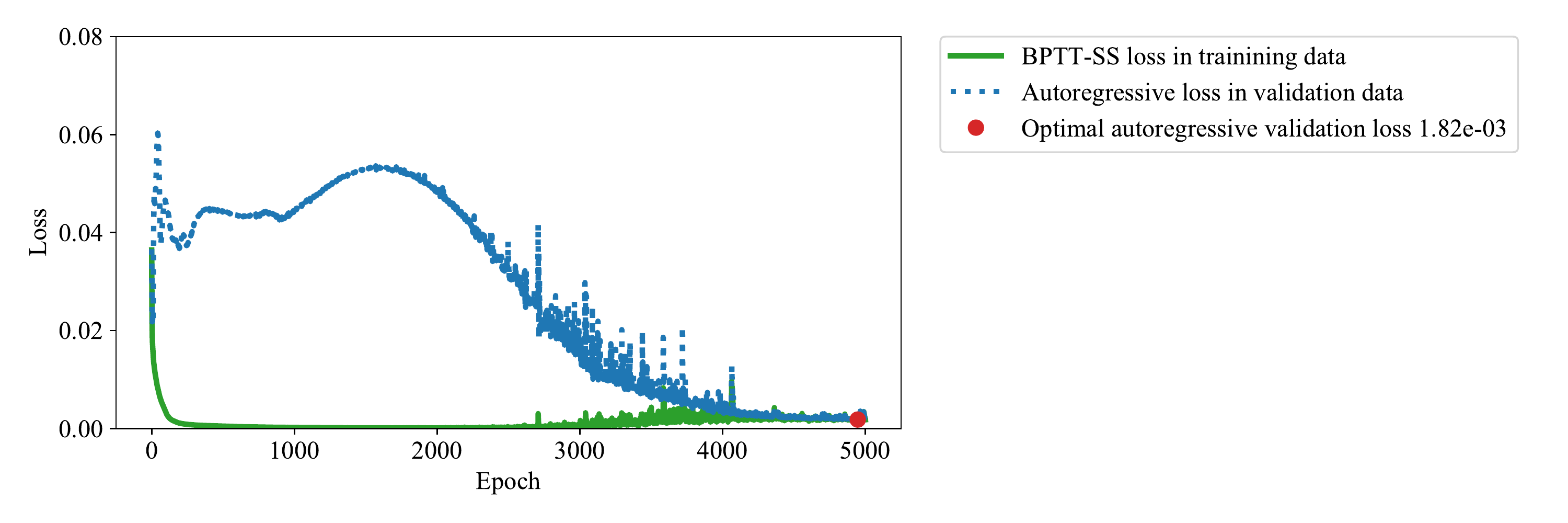}
\caption{ConvRNN trained with BPTT-SS}
\label{fig:NSC600_ConvRNN:Comparison_losses_BPTT-SS}
\end{subfigure}
\hfill 
\begin{subfigure}[tbhp]{0.75\textwidth}
\centering
\includegraphics[width=1.0\textwidth]{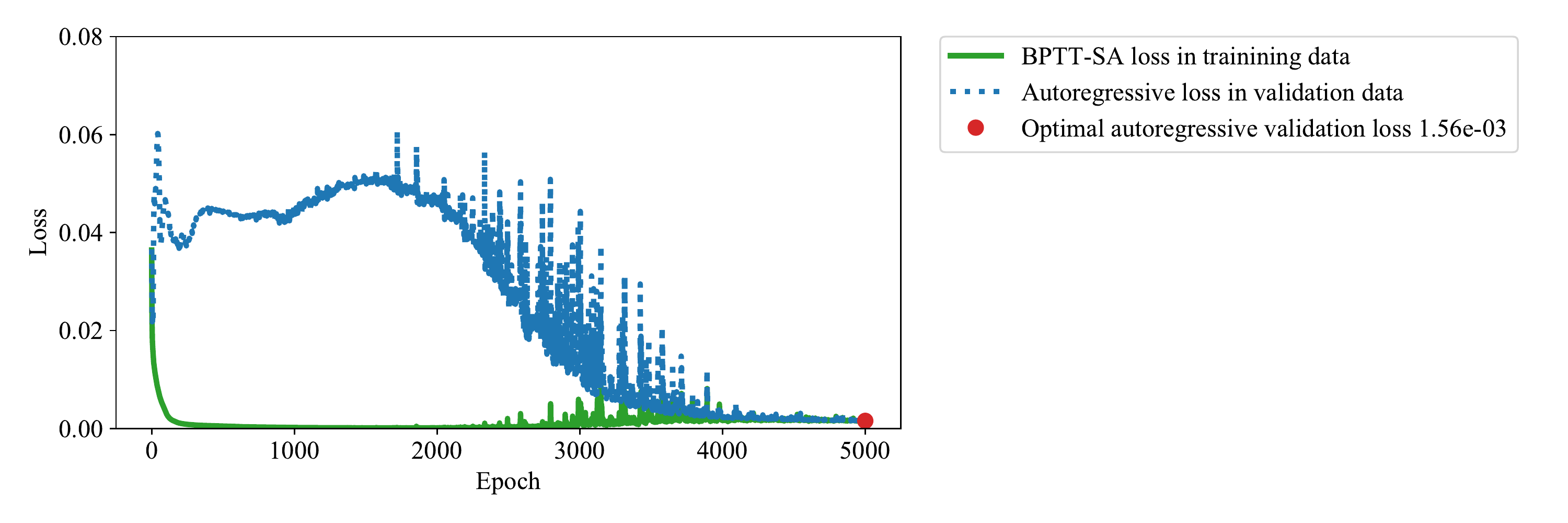}
\caption{ConvRNN trained with BPTT-SA}
\label{fig:NSC600_ConvRNN:Comparison_losses_BPTT-SA}
\end{subfigure}
\caption{
Evolution of the training and validation losses in ConvRNN training in the Navier-Stokes dataset.
}
\label{fig:results:nsc:training:convrnn}
\end{figure}

The evolution of the training and validation error in the CNN-RNN training on the Navier-Stokes dataset is given in~\Cref{fig:results:nsc:training:cnnrnn}.
In BPTT-TF the training error is decreasing, but it does not capture the long-term autoregressive prediction error.
For this reason, the autoregressive validation error is increasing, as the model is overfitting in the one-step ahead prediction error.
In BPTT-SS, the training error is encoding the autoregressive loss due to the scheduled sampling approach.
However, as the method is not backpropagating the gradients, training is hard as the iterative forecasting probability $p$ is increased and the training error is not reduced.
For this reason, the validation error also remains high.
In contrast, in the model trained with BPTT-SA, the autoregressive validation error is indeed decreasing demonstrating that the training loss and the gradient captures and encodes successfully the objective of long-term forecasting.
A similar behavior is observed in~\Cref{fig:results:nsc:training:convrnn} for the training of ConvRNN models.
\begin{figure}[pos=H]
\centering
\includegraphics[width=0.9\textwidth]{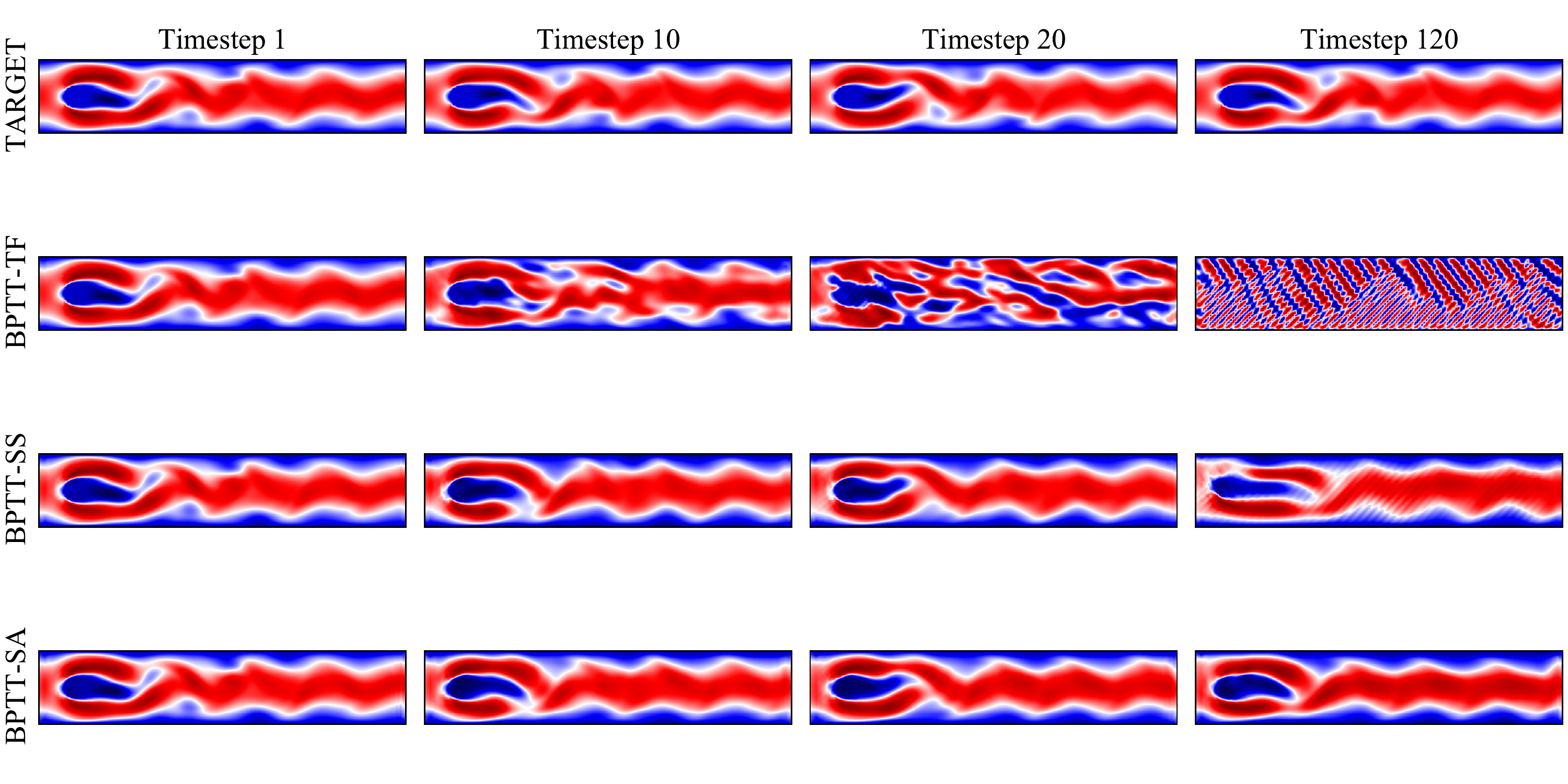}
\caption{
Prediction samples in different timesteps in the autoregressive testing mode of ConvRNN networks trained with different methods in the Navier-Stokes dataset.
}
\label{fig:NSC600_ConvRNN:ISF_IC}
\end{figure}

\begin{figure}[pos=H]
\centering
\begin{subfigure}[tbhp]{0.24\textwidth}
\centering
\includegraphics[width=1.0\textwidth,clip]{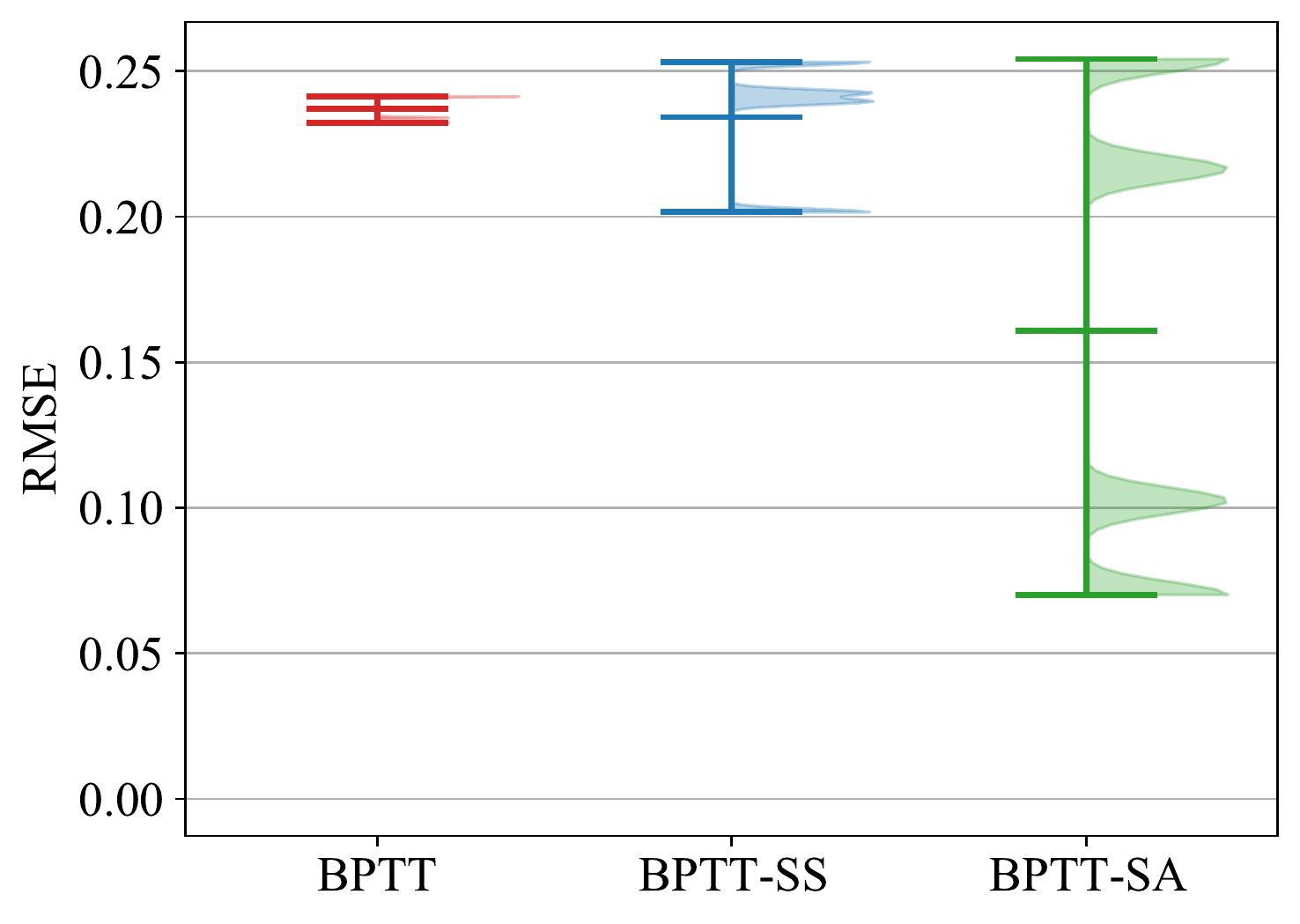}
\caption{CNN-RNN}
\label{fig:NSC600_CNNRNN:RMSE_AISF_T}
\end{subfigure}
\hfill 
\begin{subfigure}[tbhp]{0.24\textwidth}
\centering
\includegraphics[width=1.0\textwidth,clip]{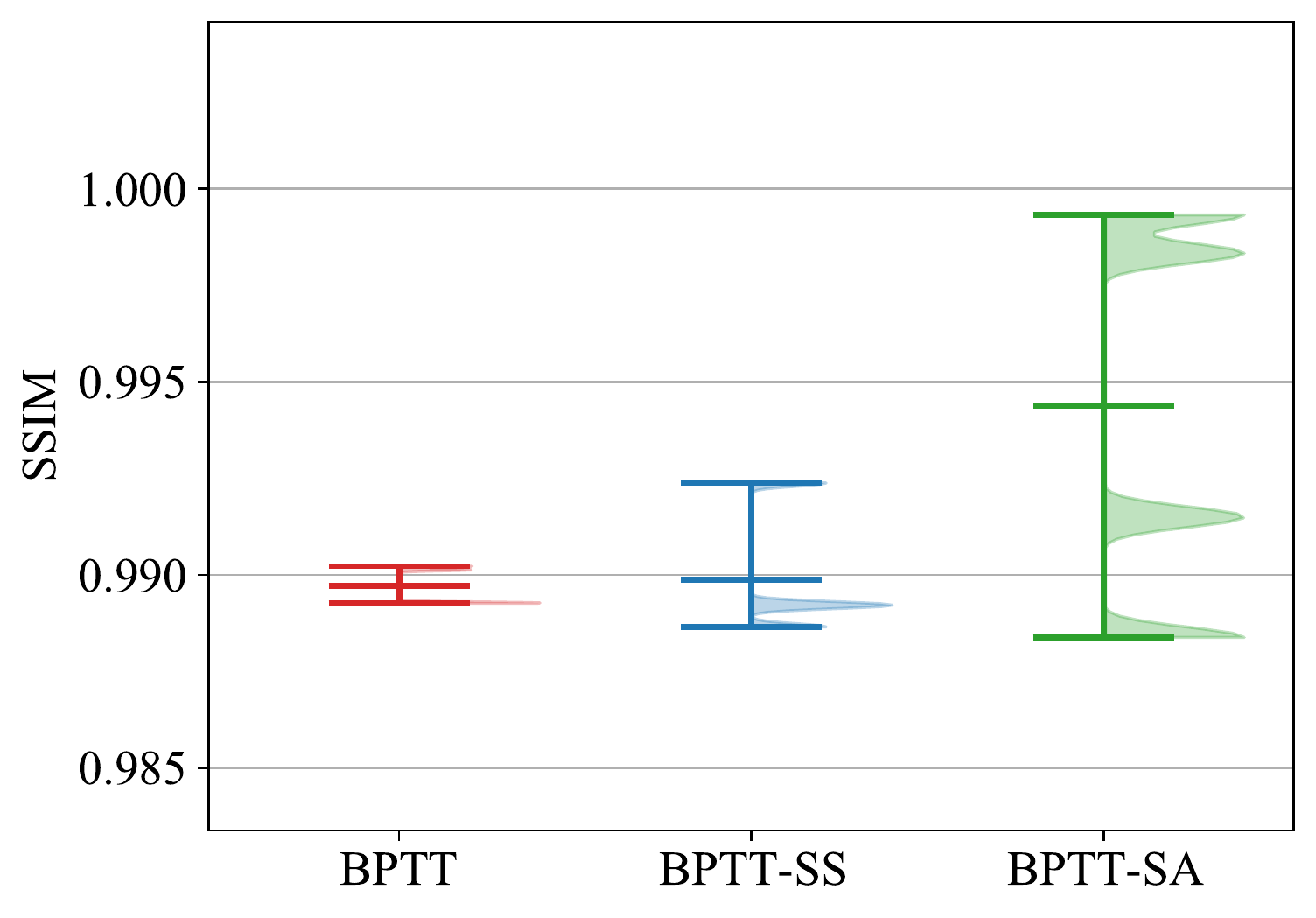}
\caption{CNN-RNN}
\label{fig:NSC600_CNNRNN:SSIM_AISF_T}
\end{subfigure}
\hfill 
\begin{subfigure}[tbhp]{0.24\textwidth}
\centering
\includegraphics[width=1.0\textwidth,clip]{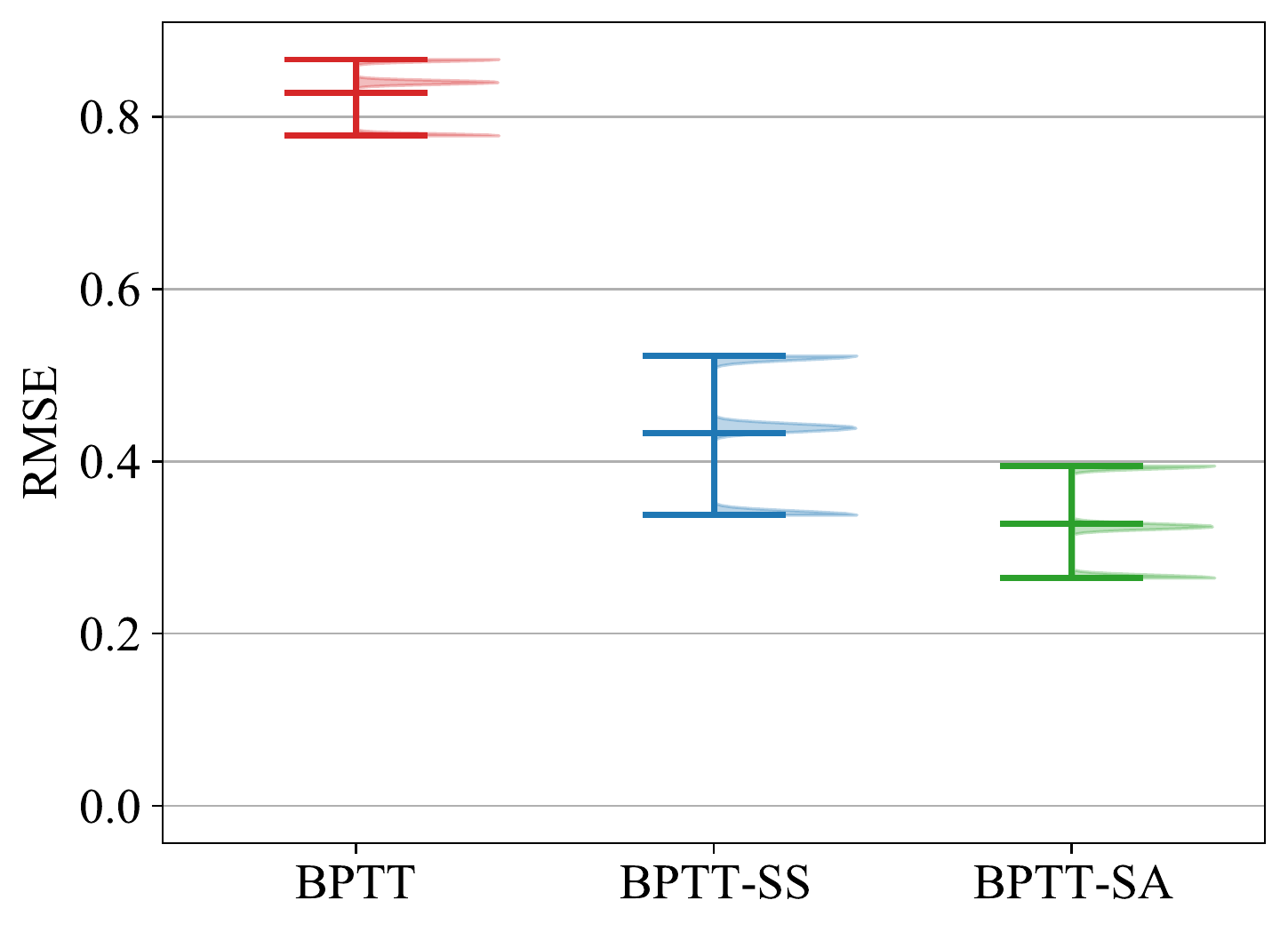}
\caption{Conv-RNN}
\label{fig:NSC600_ConvRNN:RMSE_AISF_T}
\end{subfigure}
\hfill 
\begin{subfigure}[tbhp]{0.24\textwidth}
\centering
\includegraphics[width=1.0\textwidth,clip]{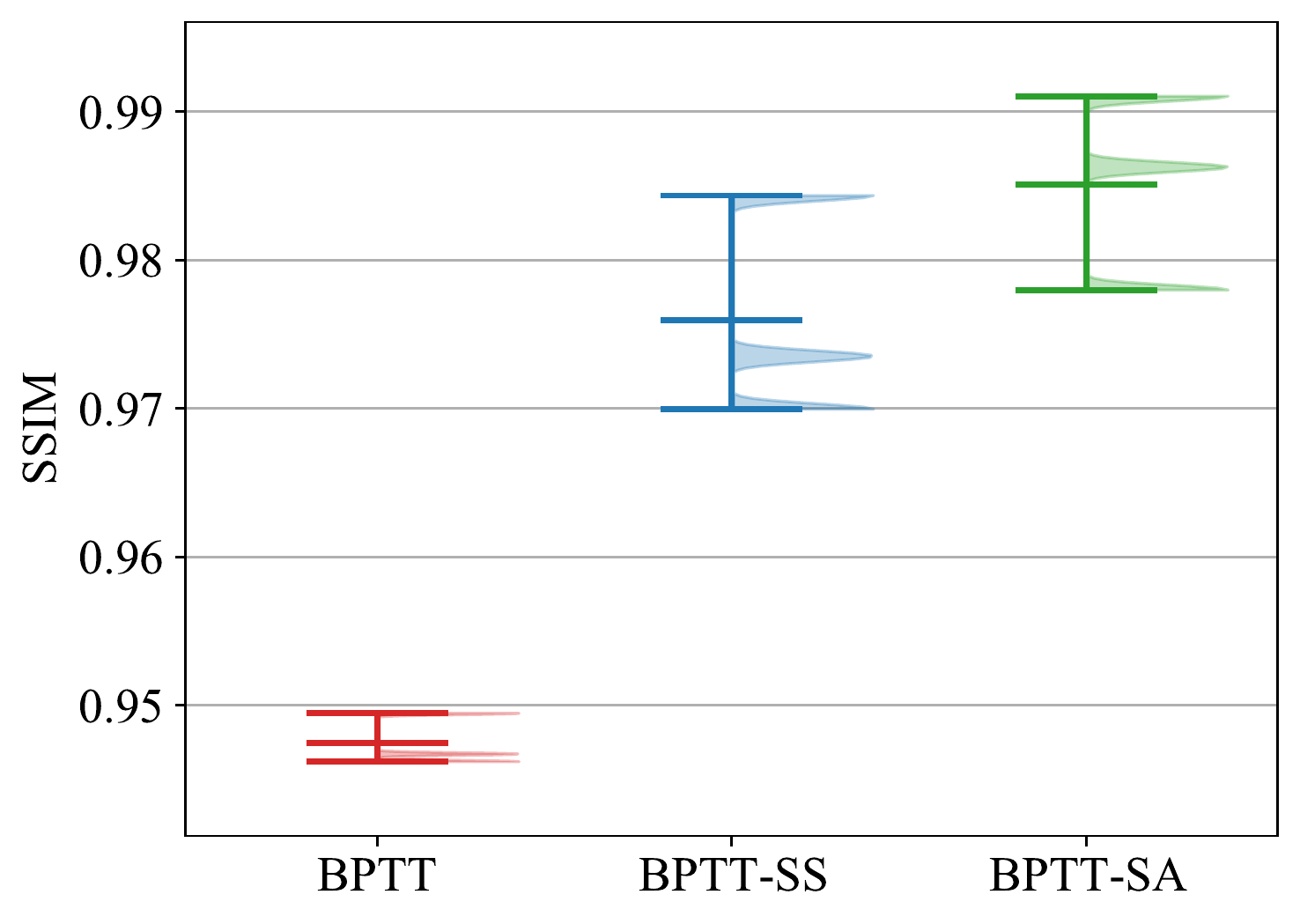}
\caption{Conv-RNN}
\label{fig:NSC600_ConvRNN:SSIM_AISF_T}
\end{subfigure}
\centering
\caption{
Evaluation of the performance of Scheduled Autoregressive BPTT (BPTT-SA) in long-term prediction on the Navier-Stokes flow past a cylinder for two model types, CNN-RNNs and ConvRNNs.
Better prediction capability of a model is demonstrated by lower RMSE and higher SSIM scores. 
}
\label{fig:results:nsc}
\end{figure}

In the autoregressive testing, we consider two comparison metrics, i.e., the RMSE error (the smaller, the better) and the structural similarity index measure (SSIM)~\citep{wang2004image} (the higher, the better).
The performance of the CNN-RNN models is illustrated in~\Cref{fig:NSC600_CNNRNN:RMSE_AISF_T,fig:NSC600_CNNRNN:SSIM_AISF_T}.
Four random seeds are considered to evaluate the robustness of the training algorithms.
BPTT-SA leads on average to a drastic reduction of the RMSE, 
and increase in the SSIM.
The same holds for ConvRNNs models as depicted in~\Cref{fig:NSC600_ConvRNN:RMSE_AISF_T,fig:NSC600_ConvRNN:SSIM_AISF_T}.
CNN-RNNs exhibit lower errors in both metrics compared to ConvRNNs as they take into account the reduced order nature of the effective dynamics, and predict on a low-dimensional latent space.

\begin{figure}[pos=H]
\begin{subfigure}[tbhp]{0.45\textwidth}
\centering
\includegraphics[width=1.0\textwidth,clip]{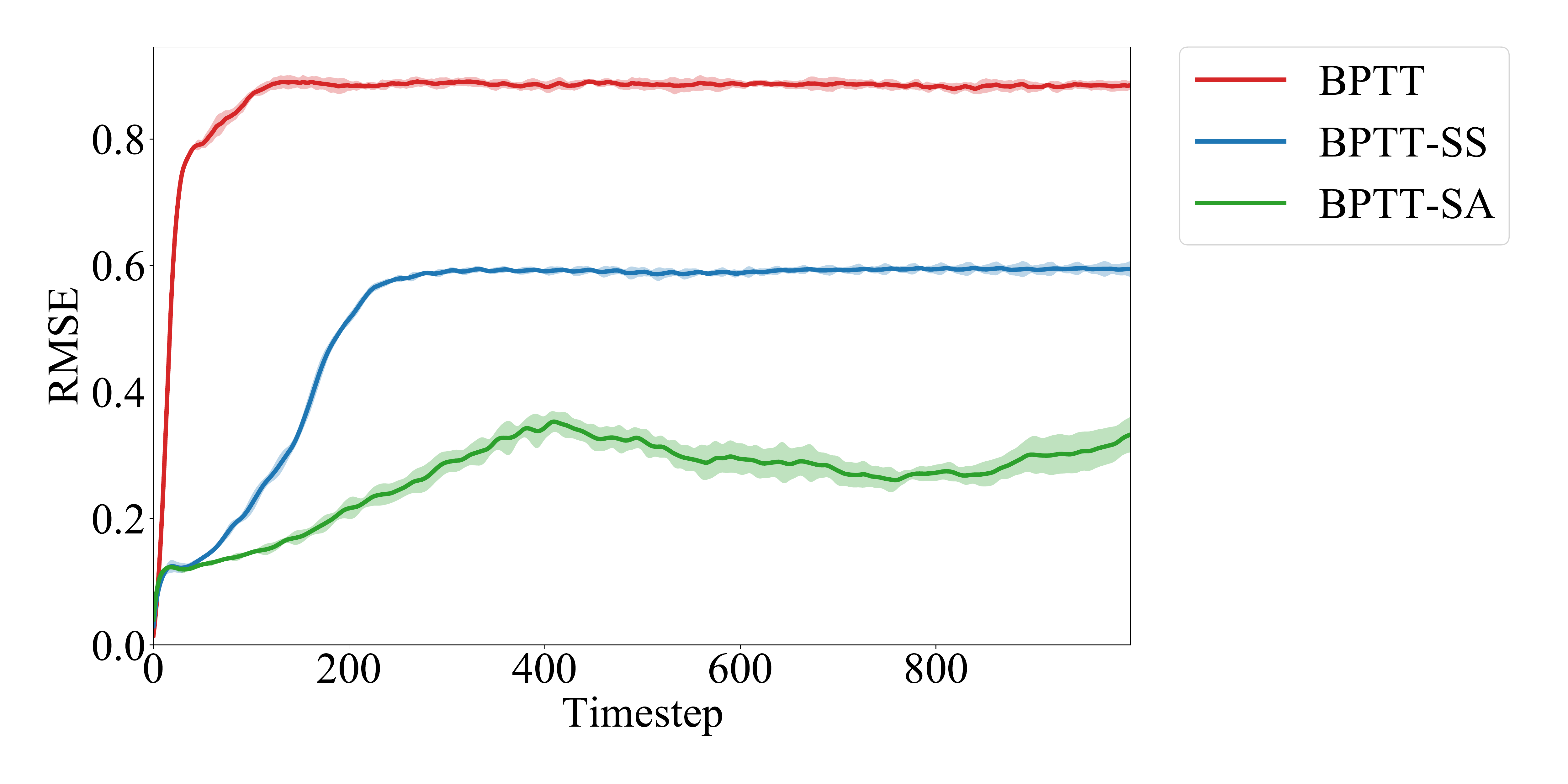}
\caption{RMSE in ConvRNNs}
\label{fig:NSC600_ConvRNN:RMSE_ISF_IC}
\end{subfigure}
\hfill 
\begin{subfigure}[tbhp]{0.45\textwidth}
\centering
\includegraphics[width=1.0\textwidth,clip]{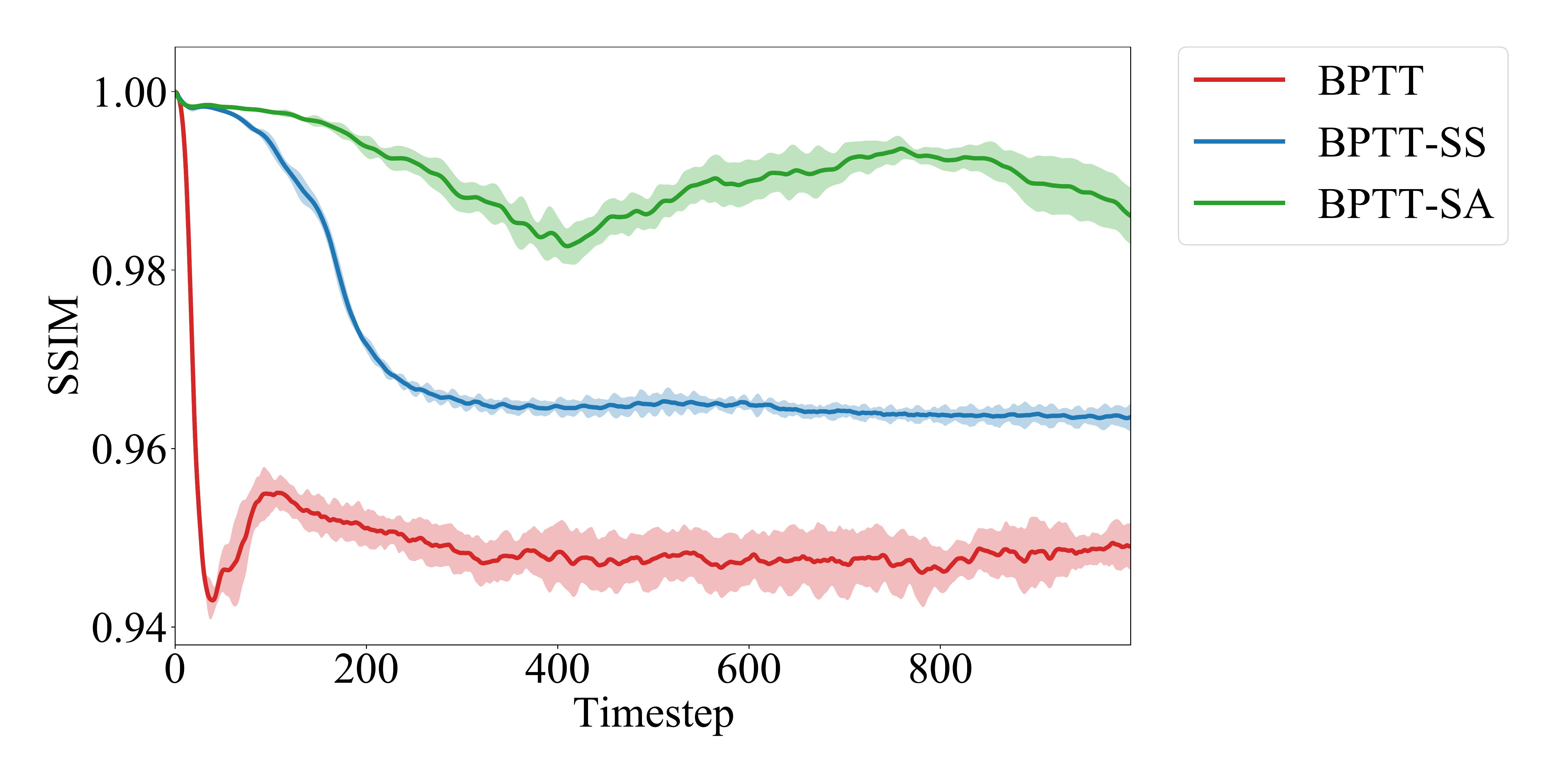}
\caption{SSIM in ConvRNNs}
\label{fig:NSC600_ConvRNN:SSIM_ISF_IC}
\end{subfigure}
\caption{
The evolution of the SSIM and RMSE errors in time in autoregressive long-term prediction in the test data on the Navier-Stokes flow past a cylinder for ConvRNNs.
Better prediction capability of a model is demonstrated by lower RMSE and higher SSIM scores. 
}
\label{fig:results:nsc:errorsintime}
\end{figure}

In~\Cref{fig:results:nsc:errorsintime}, we plot the evolution of the RMSE and the SSIM errors in time.
We observe that BPTT-SA alleviates the error propagation and leads to more accurate long-term predictions in both metrics.
In~\Cref{fig:NSC600_ConvRNN:ISF_IC}, we plot samples from the autoregressive testing phase for the ConvRNN models.
We observe that models trained with BPTT and BPTT-SS lead to unphysical predictions after some time-steps.
At lead time $T=120$, only BPTT-SA captures the flow characteristics.

\section{Discussion}
\label{sec:discussion}
We find that BPTT-SA  is particularly  useful in scenarios where creating models for long-term forecasting is necessary but creating various models for different lead times is not possible or expensive. 
The BPTT-SA method is applicable to any recurrent architecture without additional training time or memory cost. 
BPTT-SA has many potential applications, including improving the long-term prediction capabilities of data-driven surrogate/reduced order models of dynamical systems, Computational Fluid Dynamic (CFD), or Finite Element (FEM) codes, and environment dynamics models for model-based Reinforcement Learning~\citep{du2019good}. 
Additionally, the method can be used to fine-tune any recurrent architecture to achieve state-of-the-art results in long-term prediction across various applications, as demonstrated by the recent work of~\cite{su2020convolutional}. 
However, in the case of low-dimensional time series, the results of this study indicate that the merits of BPTT-SA regarding long-term forecasting performance are marginal.

\section{Summary}

In this study, we introduce a new method called scheduled autoregressive BPTT (BPTT-SA) to address the exposure bias issue in RNNs that arises in iterative forecasting. 
We compare the performance of BPTT-SA to standard BPTT and a schedule sampling approach in low dimensional time-series problems, and the Navier-Stokes flow past a cylinder. 

Our results show that BPTT-SA can effectively reduce errors in long-term, high-dimensional spatiotemporal prediction in ConvRNNs and CNN-RNNs for the Navier-Stokes flow without incurring any additional training costs. 
Future research will include the evaluation of the method in  climate~\citep{rasp2020weatherbench} and fluid flow~\citep{wiewel2019latent} datasets, exploration of alternative sampling schedules, and the study of more sophisticated sampling mechanisms, i.e., importance sampling based on the prediction error.

\printcredits

\bibliographystyle{cas-model2-names}

\bibliography{bibliography}


\bio{}
\endbio

\newpage
\appendix

\section{Lemma 1}
\label{app:sec:lemma}

\begin{lemma}
Assume that there are three sequences $a_t, b_t$ and $c_t$, with $a_0=1$ and $a_t=b_t + c_t a_{t-1}$ for $t \in {1,2,\dots, T}$.
For $t>1$ it holds that:
\begin{equation}
a_t = b_t + \sum_{i=1}^{t-1}
\Big(
\prod_{j=i+1}^{t} c_j
\Big)
b_i.
\end{equation}
\end{lemma}

\section{Hyperparameters for Darwin Sea Level Temperatures modeling}
The hyperparameters for the networks employed in the Darwin dataset are given in~\Cref{app:tab:darwin}.
\begin{table}[pos=H]
\caption{Hyperparameter tuning in Darwin dataset}
\label{app:tab:darwin}
\centering
\resizebox{0.4\textwidth}{!}{$
\begin{tabular}{ |c|c|c| } 
\hline
\text{Hyperparameter} & 
\text{Values} \\  \hline \hline
Optimizer& Adam \\
Batch size& $32$ \\
Initial learning rate & $0.0001 $ \\
Max Epochs& $2000$ \\
Random Seed& $\{1,\dots, 10\}$ \\
BPTT sequence length $L$ & $100$ \\
Prediction horizon & $ 100 $ \\
Number of testing initial conditions & $32 $ \\
Number of LSTM layers & $1$ \\
Size of LSTM layers & $100$ \\
Activation of LSTM Cell & $\operatorname{tanh}$ \\
Scaling & $[0,1]$ \\
\hline 
\end{tabular}
$}
\end{table}

\section{Hyperparameters for Mackey Glass modeling}
\label{app:mackey:hyp}
The hyperparameters for the networks employed in the Mackey Glass system are given in~\Cref{app:tab:mackey}.
\begin{table}[pos=H]
\caption{Hyperparameter tuning in Mackey Glass system}
\label{app:tab:mackey}
\centering
\resizebox{0.4\textwidth}{!}{$
\begin{tabular}{ |c|c|c| } 
\hline
\text{Hyperparameter} & 
\text{Values} \\  \hline \hline
Optimizer& Adam \\
$\operatorname{SNR}$& $\{10, 60\}$ \\
Batch size& $32$ \\
Initial learning rate & $0.0005 $ \\
Max Epochs& $2000$ \\
Random Seed& $\{1,\dots, 10\}$ \\
BPTT sequence length $L$ & $40$ \\
Prediction horizon & $ 896 $ \\
Number of testing initial conditions & $100 $ \\
Number of LSTM layers & $1$ \\
Size of LSTM layers & $100$ \\
Activation of LSTM Cell & $\operatorname{tanh}$ \\
Scaling & $[0,1]$ \\
\hline 
\end{tabular}
$}
\end{table}

\section{Hyperparameters for Navier Stokes Flow modeling}
\label{app:ns:hyp}
The hyperparameters for the networks employed in the Navier-Stokes dataset are given in~\Cref{app:tab:ns:convrnn} for ConvRNNs and in~\Cref{app:tab:ns:cnnrnns} for CNN-RNNs.
The autoencoder of CNN-RNNs is composed of consecutive Convolutional layers, Average pooling, CELU activation, and Batch-Norm layers.
The exact architecture is given in~\Cref{app:tab:ns:cnn}.
The autoencoder is reducing the dimensionality on a $\bm{z} \in \mathbb{R}^5$ latent space.
An LSTM with $24$ units is predicting on this latent space.
\begin{table}[pos=H]
\caption{Architecture of CNN of CNN-RNNs in Navier-Stokes dataset}
\label{app:tab:ns:cnn}
\centering
\resizebox{0.8\textwidth}{!}{$
\begin{tabular}{ |c|c|c| } 
\hline
\text{Layer} &
\text{Encoder} &
\text{Decoder}
 \\  \hline \hline
1 & ZeroPad2d(padding=(5, 5, 5, 5), value=0.0) & Upsample(scale\_factor=2.0, mode=bilinear)\\
2 & Conv2d(3, 5, kernel\_size=(11, 11), stride=(1, 1)) & ConvTranspose2d(1, 2, kernel\_size=(3, 3), stride=(1, 1), padding=[1, 1]) \\
3 & AvgPool2d(kernel\_size=2, stride=2, padding=0)  & CELU(alpha=1.0)\\
4 & CELU(alpha=1.0) & BatchNorm2d(2, eps=1e-05, momentum=0.1, affine=False)\\
5 & BatchNorm2d(5, eps=1e-05, momentum=0.1, affine=False) & Upsample(scale\_factor=2.0, mode=bilinear)\\
6 & ZeroPad2d(padding=(4, 4, 4, 4), value=0.0) & ConvTranspose2d(2, 20, kernel\_size=(3, 3), stride=(1, 1), padding=[1, 1])\\
7 & Conv2d(5, 10, kernel\_size=(9, 9), stride=(1, 1)) & CELU(alpha=1.0)\\
8 & AvgPool2d(kernel\_size=2, stride=2, padding=0) & BatchNorm2d(20, eps=1e-05, momentum=0.1, affine=False)\\
9 & CELU(alpha=1.0) & Upsample(scale\_factor=2.0, mode=bilinear)\\
10 & BatchNorm2d(5, eps=1e-05, momentum=0.1, affine=False) &  ConvTranspose2d(20, 10, kernel\_size=(7, 7), stride=(1, 1), padding=[3, 3])\\
11 & ZeroPad2d(padding=(3, 3, 3, 3), value=0.0) & CELU(alpha=1.0)\\
12 & Conv2d(10, 20, kernel\_size=(7, 7), stride=(1, 1)) & BatchNorm2d(10, eps=1e-05, momentum=0.1, affine=False)\\
13 & AvgPool2d(kernel\_size=2, stride=2, padding=0) &Upsample(scale\_factor=2.0, mode=bilinear) \\
14 & Conv2d(5, 10, kernel\_size=(9, 9), stride=(1, 1)) &  ConvTranspose2d(10, 5, kernel\_size=(9, 9), stride=(1, 1), padding=[4, 4]) \\
15 & BatchNorm2d(20, eps=1e-05, momentum=0.1, affine=False) &  CELU(alpha=1.0) \\
16 & ZeroPad2d(padding=(1, 1, 1, 1), value=0.0) &  BatchNorm2d(5, eps=1e-05, momentum=0.1, affine=False)\\
17 & Conv2d(20, 2, kernel\_size=(3, 3), stride=(1, 1)) &  Upsample(scale\_factor=2.0, mode=bilinear) \\
18 & AvgPool2d(kernel\_size=2, stride=2, padding=0) & ConvTranspose2d(5, 3, kernel\_size=(11, 11), stride=(1, 1), padding=[5, 5]) \\
19 & CELU(alpha=1.0) & 0.5 + 0.5 Tanh() \\
20 & BatchNorm2d(20, eps=1e-05, momentum=0.1, affine=False)  &\\
21 & ZeroPad2d(padding=(1, 1, 1, 1), value=0.0)  &\\
22 & Conv2d(20, 2, kernel\_size=(3, 3), stride=(1, 1))  &\\
23 & AvgPool2d(kernel\_size=2, stride=2, padding=0)  &\\
24 & CELU(alpha=1.0)  &\\
Latent & $\bm{z} \in \mathbb{R}^5$  &\\
Scaling & $[0,1]$ & \\
\hline 
\end{tabular}
$}
\end{table}

\begin{table}[pos=H]
\caption{Hyperparameters of CNN-RNNs in Navier-Stokes dataset}
\label{app:tab:ns:cnnrnns}
\centering
\resizebox{0.4\textwidth}{!}{$
\begin{tabular}{ |c|c|c| } 
\hline
\text{Hyperparameter} & 
\text{Values} \\  \hline \hline
Optimizer& Adam \\
Batch size& $32$ \\
Initial learning rate & $0.001 $ \\
Max Epochs& $1000$ \\
Random Seed& $\{1, 2, 3, 4\}$ \\
BPTT sequence length $L$ & $20$ \\
Prediction horizon & $ 1000 $ \\
Number of testing initial conditions & $10 $ \\
RNN Cell & LSTM \\
Number of RNN layers & $1$ \\
Size of RNN layers & $24$ \\
Scaling & $[0,1]$ \\
\hline 
\end{tabular}
$}
\end{table}

\begin{table}[pos=H]
\caption{Hyperparameters of ConvRNNs in Navier-Stokes dataset}
\label{app:tab:ns:convrnn}
\centering
\resizebox{0.4\textwidth}{!}{$
\begin{tabular}{ |c|c|c| } 
\hline
\text{Hyperparameter} & 
\text{Values} \\  \hline \hline
Optimizer& Adam \\
Batch size& $16$ \\
Initial learning rate & $0.0001 $ \\
Max Epochs& $5000$ \\
Random Seed& $\{1,2,3\}$ \\
BPTT sequence length $L$ & $50$ \\
Prediction horizon & $ 1000 $ \\
Number of testing initial conditions & $10 $ \\
RNN Cell & LSTM \\
Number of RNN layers & $1$ \\
Size of RNN layers & $8$ \\
Kernel size & $5$ \\
Scaling & $[0,1]$ \\
\hline 
\end{tabular}
$}
\end{table}

\end{document}